\documentclass[10pt,twocolumn,letterpaper]{article}
\usepackage{iccv}
\usepackage{times}
\usepackage{epsfig}
\usepackage{graphicx}
\usepackage{amsmath}
\usepackage{amssymb}
\usepackage{subfigure}
\usepackage{multirow}
\usepackage{url}

\usepackage{enumitem}
\setlist{noitemsep,topsep=0pt,parsep=0pt,partopsep=0pt,leftmargin=12pt}

\newcommand{\set}[1]{\left\{#1\right\}}

\newcommand{\X}{\mathcal{X}}
\newcommand{\vx}{\mathbf{x}}
\newcommand{\hvx}{\mathbf{\hat{x}}}
\newcommand{\hX}{\hat{\X}}

\newcommand{\reals}[1]{\mathbb{R}^{#1}}

\newcommand{\vz}{\mathbf{z}}
\newcommand{\Z}{\mathbf{Z}}
\newcommand{\pZ}{\mathcal{Z}}
\newcommand{\hZ}{\hat{\mathbf{Z}}}
\newcommand{\hvz}{\hat{\vz}}

\newcommand{\Gp}{G_p}
\newcommand{\Gr}{G_r}
\newcommand{\Gs}{G_s}

\newcommand{\name}{InSeGAN\xspace}

\newcommand{\template}{T}

\newcommand{\F}{\mathcal{F}}

\newcommand{\loss}{\mathcal{L}}
\newcommand{\expect}{\mathbb{E}}
\newcommand{\enorm}[1]{\left\|{#1}\right\|}
\newcommand{\bF}{\bar{\mathcal{F}}}
\newcommand{\lnorm}[1]{\left\|{#1}\right\|_1}
\newcommand\inv[1]{#1\!\!\raisebox{1.15ex}{$\scriptscriptstyle-\!1$}}
\newcommand{\eye}{\mathbf{I}}

\DeclareMathOperator{\normal}{N}
\DeclareMathOperator{\SO3}{SO(3)}
\DeclareMathOperator{\SE3}{SE(3)}
\DeclareMathOperator{\STN}{STN}
\DeclareMathOperator{\OT}{OT}

\DeclareMathOperator*{\argmin}{arg\,min}

\usepackage[pagebackref=true,breaklinks=true,letterpaper=true,colorlinks,bookmarks=false]{hyperref}

\iccvfinalcopy 

\def\iccvPaperID{9605} 

\ificcvfinal\fi

\begin{document}

\title{InSeGAN: A Generative Approach to Segmenting \\Identical Instances in Depth Images}

\author{Anoop Cherian$^1$\quad Gon\c{c}alo Dias Pais$^2$\thanks{Work done as part of an internship at MERL.} \quad Siddarth Jain$^1$ \quad Tim K. Marks$^1$ \quad Alan Sullivan$^1$\\
$^1$Mitsubishi Electric Research Labs (MERL), Cambridge, MA\\
$^2$Instituto Superior Técnico, University of Lisbon, Portugal\\
\small{\texttt{$^1$\{cherian, sjain, tmarks, sullivan\}@merl.com \quad $^2$goncalo.pais@tecnico.ulisboa.pt}}
}

\maketitle
\ificcvfinal\thispagestyle{empty}\fi

\begin{abstract}
\vspace{-5pt}
In this paper, we present \emph{\name}, an unsupervised 3D generative adversarial network (GAN) for segmenting (nearly) identical instances of rigid objects in depth images.  Using an analysis-by-synthesis approach, we design a novel GAN architecture to synthesize a multiple-instance depth image with independent control over each instance. \name takes in a set of code vectors (e.g., random noise vectors), each encoding the 3D pose of an object that is represented by a learned implicit object template. The generator has two distinct modules. The first module, the instance feature generator, uses each encoded pose to transform the implicit template into a feature map representation of each object instance. The second module, the depth image renderer, aggregates all of the single-instance feature maps output by the first module and generates a multiple-instance depth image.  
A discriminator distinguishes the generated multiple-instance depth images from the distribution of true depth images. 

To use our model for instance segmentation, we propose an instance pose encoder that learns to take in a generated depth image and reproduce the pose code vectors for all of the object instances. To evaluate our approach, we introduce a new synthetic dataset, ``Insta-10'', consisting of 100,000 depth images, each with 5 instances of an object from one of 10 classes. Our experiments on Insta-10, as well as on real-world noisy depth images, show that \mbox{\name} achieves state-of-the-art performance, often outperforming prior methods by large margins. Code and data for the approach are available at \url{https://www.merl.com/research/license/InSeGAN#download}
\end{abstract}
\vspace{-14pt}

\section{Introduction}
\vspace{-6pt}
\label{sec:intro}
\begin{figure}[!h]
    \centering
    \includegraphics[width=8cm,trim={3.5cm 5cm 9.5cm 5.3cm},clip]{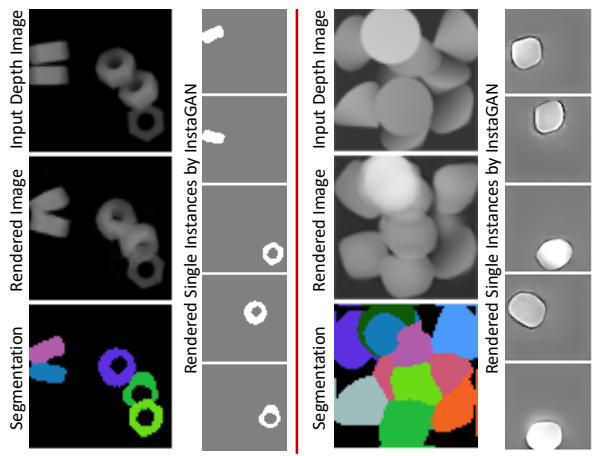}
    \caption{Segmentations and single instances disentangled by \name on two multiple-instance depth images (\emph{Left:} Nut with 5 instances;~\emph{Right:} Cone with 10 instances---challenging). \name needs only unlabelled multiple-instance depth images for training. For each input image, the hallucinated depth image (``rendered image'') and the single instances disentangled from the depth image (``rendered single instances'') are shown. We use depth pooling (Z-buffering) and thresholding to produce instance segmentation (``segmentation'') from the generated single instances. Note that our method learns the shape of the object automatically.}
    \label{fig:first_page}
    \vspace*{-0.5cm}
\end{figure}
Identifying (nearly) identical instances of objects is a problem that is ubiquitous in daily life. For example, when taking a paperclip from a container, choosing an apple from a box, or removing a book from a library shelf, humans subconsciously solve this problem because we have an understanding of what the individual instances are. However, when robots are deployed for such a \emph{picking} task, they need to be able to identify the instances for planning their grasp and approach~\cite{mahler2017learning,alonso2018current}. Such a problem is commonplace in large manufacturing, industrial, and agricultural contexts~\cite{xie2020unseen,xiang2020learning,wu2020deep,hu2019segmentation,jafari2018ipose}. Examples include an industrial robot picking parts from a bin, a warehouse robot picking and placing packages into a delivery truck, or even a fruit-picking robot picking identical fruits in a supermarket. In these scenarios, the robot's owners often have no access to a 3D model of the object to be picked, and annotating individual instances for training can be costly, inconvenient, and unscalable. However, they may have access to a large number of unlabeled images each containing multiple instances of the object, such as depth images of boxes as they travel on a conveyor belt from production to a packaging section. Our goal in this paper is to build an unsupervised instance segmentation algorithm using unlabeled depth images, each containing multiple identical instances of a 3D object.

Our problem setting is very different from the instance segmentation setups that are typically considered, such as that of Mask-RCNN~\cite{he2017mask}, 3D point cloud segmentation~\cite{liu2018path}, scene understanding~\cite{greff2019multi}, and others~\cite{hafiz2020survey,johnson2017clevr,lempitsky2010learning}. While these methods usually consider segmenting instances from cluttered backgrounds, our backgrounds are usually simple; however, the foreground instances can be heavily (self-)occluded or may vary drastically in appearance across their poses (see Fig.~\ref{fig:first_page} for example). Prior methods to solve our instance segmentation problem use 3D CAD models~\cite{ikeuchi1987generating}, fit the 3D instances using primitive shapes~\cite{harada2013probabilistic}, or use classical image-matching techniques to identify the instances~\cite{buchholz2015bin,piccinini2013sift}. More recently, some have attempted to solve this task using deep learning approaches. For example, in Wu et al.~\cite{wu2019unsupervised}, a 3D rendering framework is presented that is trained to infer the segmentation masks; however, their losses are prone to local minima. In the recent IODINE~\cite{greff2019multi}, MONET~\cite{burgess2019monet}, and Slot Attention~\cite{locatello2020object} deep models, the focus is on RGB scene decomposition, and may not generalize to segmenting foreground instances from each other. 

In this paper, we present a general unsupervised framework for instance segmentation in depth images, which we call \emph{\name}. Our model is inspired by a key observation made in several recent works (e.g.,~\cite{karras2019style,nguyen2019hologan}) that random noise that is systematically injected into a generative adversarial network (GAN) can control various attributes of the generated images. A natural question then is whether we can generate an image with a specific number of instances feeding in the respective number of random noise vectors. If so, then instance segmentation could be reduced to simply decoding a test image into several noise vectors, each of which generates its respective instance. \name implements this idea using a combination of a 3D GAN and an image encoder within an \emph{analysis-by-synthesis} framework, illustrated in Fig.~\ref{fig:main}. The training data consist of an unlabeled collection of depth images, each image consisting of $n$ instances of a rigid object. \name learns an implicit 3D representation of the object shape and a pose decoder that maps random noise vectors to 3D rigid transformations. The generator has two stages. In the first stage, the decoded 3D transformation is applied to the implicit object template, which an \emph{instance feature generator} converts into a feature-map representation of a single object instance. After the first stage generates $n$ such instance representations from $n$ random noise vectors, the second stage aggregates these instance representations and feeds them into a \emph{depth image renderer} to produce synthetic depth images that are similar in distribution to the training images, as enforced via a discriminator. To achieve instance segmentation, we train an encoder that takes as input a generated multiple-instance depth image and encodes it into a latent space in which it must match the random noise vectors that originally generated the images in the GAN stream, thus closing the generation cycle. At inference time, a given depth image first goes through the encoder to get its set of single-instance latent vectors; these are then fed into the GAN to synthesize each instance (each image segment) individually. Results on two example test images are shown in Fig.~\ref{fig:first_page}.

While the task of instance segmentation has been approached in various contexts, there is no existing dataset that encompasses this task in the context we are after in this paper. For example, images in standard datasets such as MSCOCO~\cite{lin2014microsoft} and CityScapes~\cite{cordts2016cityscapes} contain objects of several different classes and background, which may not belong to a common latent space. We introduce a new dataset, dubbed ``Insta-10,'' consisting of 10 object classes and 10,000 depth images per class. Each image was rendered using a physics engine that simulated a bin into which 5 instances of an object are randomly dropped, resulting in arbitrary poses of the objects in the rendered depth images. The instances can have significant occlusions and size variations (due to varying distances from the camera), making the task very challenging. We use this dataset to compare our scheme with closely related methods. We also apply our instance segmentation approach to a real-world dataset of blocks in noisy depth images.  Our results show that \name outperforms all of the prior methods by a significant margin on most of the object classes.

We now summarize this paper's primary contributions:
\begin{itemize}
    \item We propose \name, a 3D GAN that learns to generate multiple-instance depth images from sets of random noise vectors in an unsupervised manner. 
    
    \item We propose a two-stage generator structure for \name, in which the first stage generates a feature map representation of each instance, and the second aggregates these single-instance feature maps and renders a multiple-instance depth image.
 
    \item To enable segmentation, we propose an instance pose encoder that encodes a multiple-instance depth image into a set of latent vectors that would generate it. To train this encoder, we introduce novel cycle-consistency losses. 
    
    \item We have created a new large-scale and challenging dataset, \emph{Insta-10}, which we are making public to advance research on this topic. 
    
    \item Our experiments on synthetic and real datasets demonstrate that \name achieves state-of-the-art performance. On the Insta-10 dataset, \name shows a relative improvement of nearly 35\% against the recent method of Wu et al.~\cite{wu2019unsupervised} and nearly 9.3\% against Locatello et al.~\cite{locatello2020object}.
\end{itemize}

\section{Related Work}
\vspace{-6pt}
\label{sec:related_work}
In this section, we review some of the closely related approaches to our method.

\noindent{\bf Multiple Objects and Instance Segmentation:}
In \mbox{IODINE}~\cite{greff2019multi}, a variational generative model is proposed for instance segmentation of RGB images using an iterative refinement of latent vectors to characterize the object instances, similar to an expectation maximization (EM) algorithm. Their key idea is to use a fixed number of latent vectors to describe the scene and iteratively infer the association of these vectors to the instances, an approach that can be unstable for complicated scenes (such as the depth images we consider in our dataset). In Slot Attention~\cite{locatello2020object}, abstract scene components, called slots, are learned for each instance in an unsupervised manner, but they do not account for the 3D structure of the scene or the instances. In Liao et al.~\cite{liao2020towards} and O3V-voxel~\cite{henderson2020unsupervised}, multiple object instances are created in an adversarial setting through image composition. Both of these methods produce a 3D feature latent space---the former a 2D primitive of the 3D object and the latter a 3D voxel representation---for each object instance. Using a fixed number of instances,~\cite{liao2020towards} composes the scene by projecting the primitive to create depth and alpha maps. In~\cite{henderson2020unsupervised}, the authors propose a scheme to generate a video sequence to extract the multiple instance images. They follow a framework similar to~\cite{burgess2019monet,greff2019multi}, where the initial image is generated from a sequence of real images, through an encoder. However, they generate a feature voxel representation for each object. At each time instance, each object is rendered and they are composed together. 

There are prior approaches that tackle the multiple instance segmentation problem for 2D and 3D images in a \emph{supervised} manner. Most of these methods, e.g., ~\cite{he2017mask, pinheiro2015learning}, first extract Regions of Interest (RoI) from the input, subsequently classifying the object in each selected region. Mask RCNN~\cite{he2017mask} expands Faster RCNN~\cite{ren2015faster} by creating a new segmentation branch to classify per-pixel object segments. DeepMask~\cite{pinheiro2015learning} learns those RoIs and their underlying masks, which are then passed through the Fast RCNN~\cite{girshick2015fast} for classification. Along similar lines, point cloud segmentation has been explored in several recent works. For example,~\cite{yi2019gspn,wang2018sgpn} propose a 2D architecture. GsPN~\cite{yi2019gspn} proposes a network to generate shapes with their specific segmentations and bounding boxes. SGPN~\cite{wang2018sgpn} generates a similarity matrix and group proposals to create independent clusters for classification. In contrast to these popular methods for instance segmentation, we differ in that we approach the problem from an unsupervised perspective.

\noindent{\bf 3D Disentanglement:} Several recent works have proposed approaches for disentangling 3D attributes using deep learning via implicit or explicit representations. Deep Voxels~\cite{sitzmann2019deepvoxels} proposes a synthesis approach to learning an implicit 3D representation of the object. The method learns to synthesize novel perspectives of an object from a learned voxel feature volume. From these voxels, one may create an explicit 3D model of the object. However, their model is not generative and requires camera parameters. HoloGAN~\cite{nguyen2019hologan} proposes a generative method that creates an implicit 3D volume of single instances. It first learns a 3D representation, which it transforms using a target pose, then projects to 2D features and renders to a final image. Our method is inspired by HoloGAN, but we go beyond it by deriving a scheme for disentangling the object instances. Another related work is PlatonicGAN~\cite{henzler2019escaping}, which creates a 3D representation of an object while generating different unseen views via adversarial learning. However, as in HoloGAN, this method is limited to a single rotated object. 

The prior work that is most similar to ours is Wu et al.~\cite{wu2019unsupervised}, which proposes to disentangle object instances and their 6D poses in an unsupervised manner, concurrently learning an explicit 3D point cloud template of the object. While our objective is similar, our proposed framework is completely different. The framework of~\cite{wu2019unsupervised} requires explicit modeling of point occlusions and computes point cloud alignments using Chamfer distance, which make the scheme computationally expensive. We avoid these challenges by using depth images, and we introduce a discriminator that implicitly learns these steps efficiently.

\section{Proposed Method}
\vspace{-6pt}
\label{sec:proposed_method}

\begin{figure*}[ht]
\centering
\includegraphics[width=16cm,trim={0.5cm 3.5cm 0.5cm 0.8cm},clip]{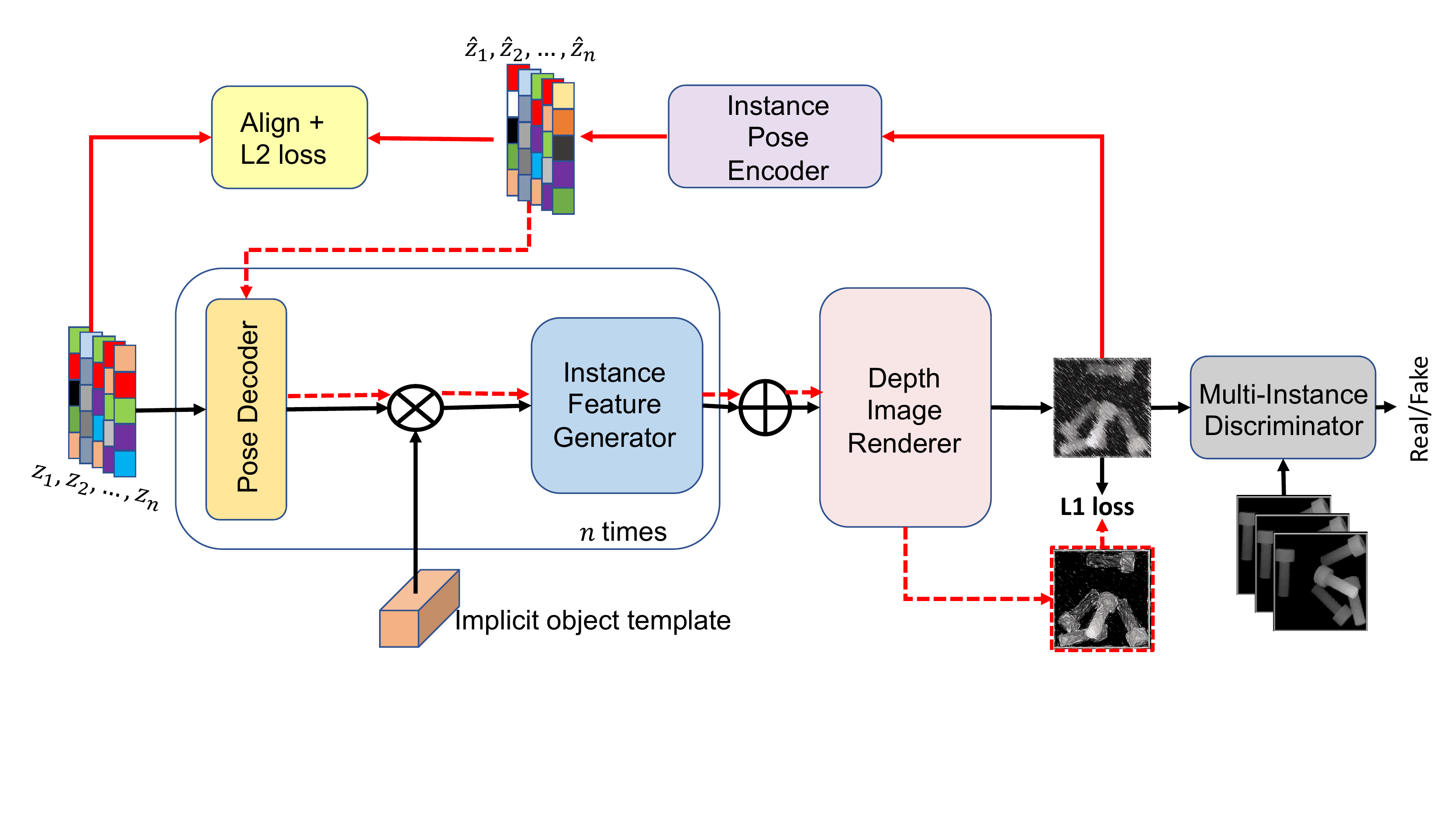}
\caption{A schematic illustration of the training scheme in \name. There are three distinct control flows in the framework, as denoted by the black, solid red, and dashed red arrows. The black arrows capture the generative process producing a multiple instance depth image, while the solid red arrows depict the scheme to encode a generated depth image to its instances. The dashed red arrows depict the control flow to train the Instance Encoder via using the encoded latent vectors to re-create the already-generated image.}\vspace{-12pt}
\label{fig:main}
\end{figure*}

Let $\X$ be a given dataset, where each $\vx \in \X$ is a depth image consisting of $n$ instances of a rigid object. Note that the same rigid object is depicted in all of the images in $\X$. To simplify the notation, we will use $\X$ to also characterize the distribution of $\vx$. Further, for clarity of presentation, we assume that $n$ is known and fixed for $\X$, however note that it is straightforward to extend InSeGAN for an arbitrary number of instances by using training images with varying numbers of instances (see Supplementary materials for details). Our goal in \name is to learn a model only from $\X$ (without any labels) such that at test time, when given a depth image $\vx$, the learned model outputs the segmentation masks associated with each instance in the depth image. In the next section, we provide a brief overview of the \name architecture, followed by a detailed look into each of its components. 

\subsection{\name Overview}
\vspace{-6pt}
The basic architecture of \name follows a standard generative adversarial framework, however with several non-trivial twists. It consists of a generator module $G$ that---instead of taking a single noise vector as input (as in a typical GAN)---takes $n$ noise vectors, $\set{\vz_1, \vz_2, \cdots, \vz_n}$, each $\vz\in\reals{d}\sim\normal(\boldsymbol{0},\eye_d)$, and generates a multiple-instance depth image as output. Thus, $G:\reals{d\times n}\to \hX$, where $\hX$ is used to signify the distribution of the generated depth images, with the limit $\hX\to\X$ when $G$ is well-trained. We denote the set of noise vectors by the matrix $\Z\in\reals{d\times n}$ and the distribution of $\Z$ as $\pZ=\left\{\normal(\boldsymbol{0},\eye_d)\right\}^n$. Next, a discriminator module $D$ is trained to distinguish whether its input is an image generated by $G$ or a sample from the data distribution $\X$. The modules $G$ and $D$ are trained in a min-max adversarial game so that $G$ learns to generate images that can fool $D$, while $D$ in turn learns to distinguish whether its inputs are real or fake; the optimum occurs when $D$ cannot recognize whether its input is from $G$ or $\X$. Apart from the generator and discriminator modules, we also have an instance pose encoder module, $E$, that is key to achieving instance segmentation. Specifically, $E:\hX\to\reals{d\times n}$ takes as input a generated depth image, and learns to output vectors that match the latent noise vectors that generated the input depth image. The essence of \name is to have the generator $G$ produce depth images for which the instance segments are implicitly known (through $\Z$), so that $E$ can be trained on them to learn to disentangle the instances. In the limit as $\hX\to\X$, as guided by the discriminator $D$, the encoder $E$ will eventually learn to do instance segmentation on real images from $\X$. An overview of the \name training pipeline is shown in Fig.~\ref{fig:main}. Next, we will describe each of the modules in detail.

\subsection{\name Generator}
\vspace{-6pt}
The key to \name is to have the generator $G$ accomplish two tasks jointly: (i) to produce depth images $\hvx$ that match the input image distribution $\X$, and (ii) to identify each object instance in the generated image $\hvx$. To this end, we note that \emph{sans} the other instances, each instance is an independent depth rendering of an object in an arbitrary 3D pose. A multiple-instance depth image may be generated by merging the individual instances, followed by depth-based inter-object occlusion reasoning.

Motivated by the above insight, we propose to separate the generator $G$ into two distinct modules: (i) an \emph{instance feature generator} that generates feature maps for single object instances, and (ii) a \emph{depth image renderer} module that aggregates the single-instance feature maps and renders the multiple-instance depth image. As the instances are assumed to be of the same object, we propose to sample each noise vector $\vz\in\Z$ from the same latent distribution, $\vz\sim\normal(\boldsymbol{0},\eye_d)$. Further, our system learns an implicit 3D object model (template) that, when geometrically transformed, produces the varied appearances of the instances.

Our first step in the generator pipeline is to produce 6-DOF (6 degrees of freedom) 3D rigid geometric transforms that can be applied to the implicit object template to produce a transformed implicit model representing each instance. To this end, each noise vector $\vz\in\Z$ is converted to an element of the special Euclidean group $\big(\!\SE3\!\big)$ using a \emph{pose decoder} module (see Fig.~\ref{fig:main}), which is a fully connected neural network that is denoted $\Gp:\reals{d}\to \reals{6}$. Given a noise vector $\vz$, $\Gp$ produces a corresponding axis-angle representation; this is next converted to an element in the Special Euclidean group, $\SE3$. We denote this operator by $\Lambda:\reals{6}\to\SO3\times\reals{3}$, i.e., $\Lambda$ produces a rotation matrix $R\in\SO3$ (the special orthogonal group) and a translation vector $t\in\reals{3}$.  A natural question in this context is why we do not sample the transformation matrix directly (as in, e.g., HoloGAN~\cite{nguyen2019hologan}). This is because, as will be clear shortly, we need to match the output of the instance pose encoder module $E$ with the pose representations of the instances, and having a Euclidean embedding for these representations offers  computationally more efficient similarity measures than directly using a rotation matrix (or axis-angle) parameterization of the underlying nonlinear geometric manifold~\cite{huynh2009metrics,zhou2019continuity}. 

Next, we use the transformation matrix thus created, i.e., $\Lambda(\Gp(\vz))$, to geometrically transform an implicit shape tensor $\template\in\reals{h\times h\times h\times k}$ (we use $h$=4, $k$=128); this parameter tensor is shared by all the instances and will, when trained (with the other modules in the pipeline), implicitly capture the shape and appearance of the object. Similar to HoloGAN~\cite{nguyen2019hologan}, we use a Spatial Transformer Network (STN)~\cite{jaderberg2015spatial} to apply the geometric transform to this implicit template. The transformed $\template$ is reshaped to $\reals{kh\times h\times h}$ and projected from 3D to 2D using a single-instance projection module, $\Gs$, to output $\hvx_f\in\reals{c\times h\times h}$, which captures the feature map representation of an instance. The above steps can be formally written as:
\begin{equation}
    \F(\vz) :=  \Gs\Bigl(\STN\bigl(\Lambda\left(\Gp\left(\vz\right)\bigr), \template\right)\Bigr).
    \label{eq:single}
\end{equation}
Next, we propose to combine these feature maps by average-pooling them, then render a multiple-instance depth image using a rendering module $\Gr$, as follows:
\begin{equation}
    \hvx = G(\Z) := \Gr(\bF) \text{ where } \bF =\frac{1}{|\Z|}\sum_{\vz\in\Z} \F(\vz),
    \label{eq:generator}
\end{equation}
where $\hvx$ denotes a depth image generated by $G$. This generative control flow is depicted using black arrows in Fig.~\ref{fig:main}.   

\subsection{\name Discriminator}
\vspace{-6pt}
As in standard GANs, the task of the discriminator $D$ is to decide whether its input comes from the natural distribution of multiple-instance depth images that produced the training set (i.e.,  $\X$) or is synthesized by our generator $G$ (i.e., $\hX$). Following standard architectures, $D$ consists of several 2D convolution, instance normalization, and LeakyRELU layers, and outputs a classification score in $[0,1]$. The objectives for training the discriminator and generator, respectively, are to minimize the following losses:
\begin{align}
    \loss_D &:= -\expect_{\vx\sim\X}\log(D(\vx)) - \expect_{\Z\sim\pZ}\log\left(1-D(G(\Z)\right),  \notag\\
    \loss_G &:= -\expect_{\Z\sim\pZ}\log D(G(\Z)).
\end{align}
The task for our discriminator is significantly different from that in prior works, as it must learn to: (i) count whether the number of rendered instances matches the number of instances in the data distribution, (ii) verify whether the rendered 3D posed objects obtained via transforming the \emph{still-being-learned} object template $T$ capture the individual appearances (which are also being learned) of the instances, and (iii) whether the layout of the rendered image is similar to the compositions of the instances in the training depth images. Fortunately, with access to a suitable dataset, $D$ can automatically achieve these desired behaviors when adversarially trained with the generator.

\begin{figure*}[ht]
\centering
\includegraphics[width=16cm,trim={0.1cm 4.0cm 0.5cm 4.5cm},clip]{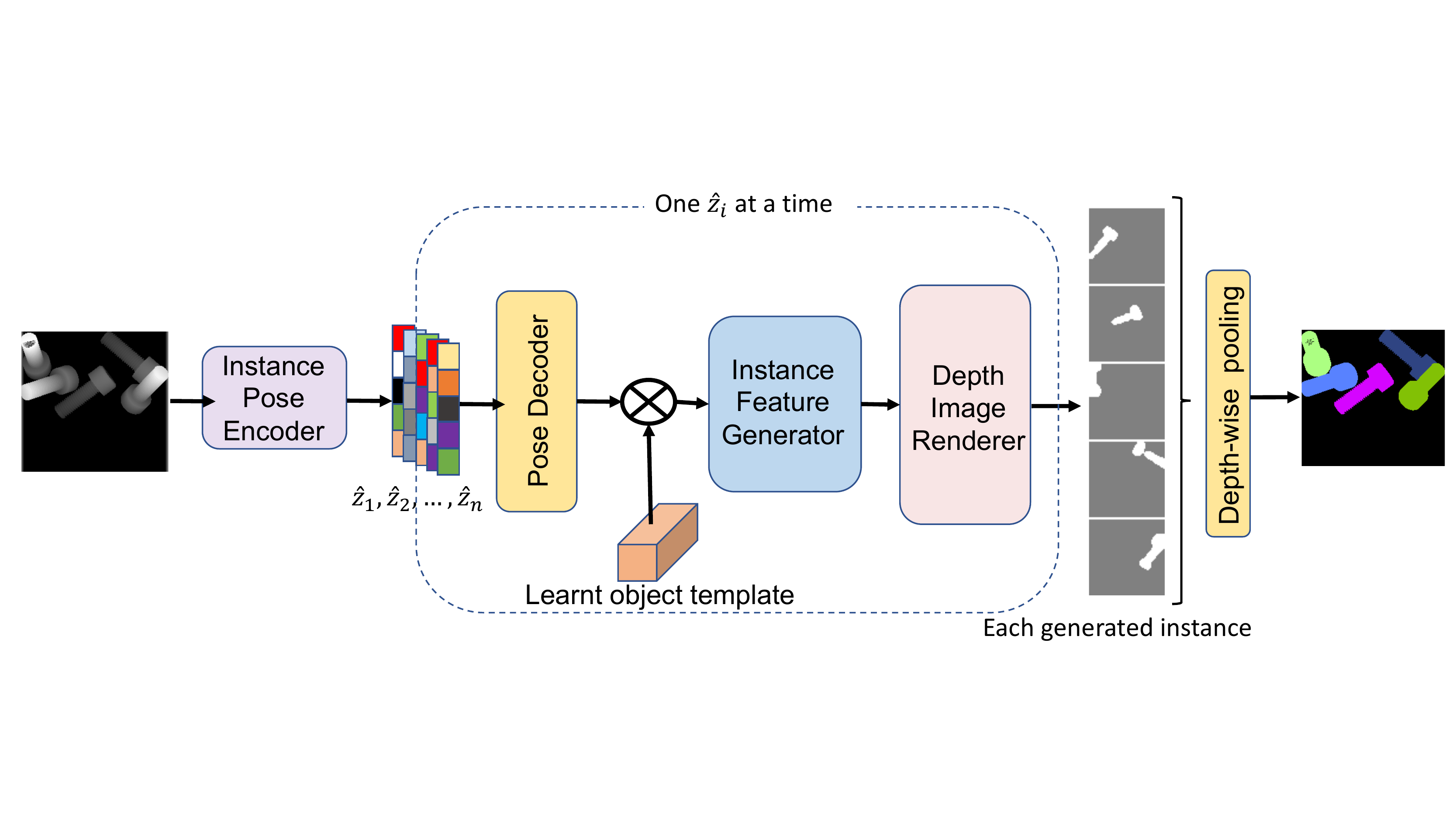}
\caption{\name inference pipeline (see Sec.~\ref{sec:inference} for details).}
\vspace{-12pt}
\label{fig:inference}
\end{figure*}

\subsection{\name Instance Pose Encoder}
\vspace{-6pt}
We now introduce our \emph{instance pose encoder} module, $E$, which is the key to instance segmentation. The task of this module is to take as input a multiple-instance depth image $\hvx$, produced by $G$, and reconstruct each of the noise vectors in $\Z$ (encoding the instance poses) that were used to generate $\hvx$. The encoder outputs $\hZ$, a set of latent vectors. Indeed, as $\hvx$ is produced by aggregating $n$ independently sampled instance appearances of the object, inverting the process amounts to disentangling $\hvx$ into its respective instances. Thus, when the generator is trained well, i.e., $\hvx\approx\vx$, we will eventually learn to disentangle each instance in a ground truth image. While this idea is conceptually simple, implementing it practically is not straightforward. There are four main difficulties: (a) the input $\Z$ to the generator and the output $\hZ$ of $E$ are unordered sets, which need to be aligned before comparing them; (b) the average pooling operator in~\eqref{eq:generator} aggregates several feature maps into one---an operation that loses the distinctiveness of each of the instance feature maps; (c) the depth renderer $\Gr$ may remove occluded parts of the instances, thus posing ambiguities when mapping them back to the noise vectors; and (d) the pose encoder $\Gp$ projects its noise input to the space of rigid body transforms, an operation that is inherently low-rank and nonlinear. We tackle these challenges via imposing losses on the encoder so that it learns to invert each module in the generator. We decompose the encoder as $E=\inv{\Gs}\circ\inv{\Gr}$, consisting of: (i) an image derenderer $\inv{\Gr}$ that takes a depth image and produces feature maps, and (ii) an instance decoder $\inv{\Gs}$ that takes the feature maps from $\inv{\Gr}$ and produces $\hZ$.

\noindent{\textbf{Alignment and Reconstruction:}} To tackle our first difficulty, (a), we propose to align the sets $\Z$ and $\hZ$ before computing a reconstruction loss on them. Specifically, we seek to find an alignment matrix $\pi\in\Pi(\Z, \hZ)$, where $\Pi$ denotes the set of all such alignments (i.e., permutations) on its inputs, such that the reconstruction loss is minimized:
\begin{equation}
    \loss_E^a \!=\! \bigl \|\Z - \pi^*(\hZ) \bigr\|^2 \!, \text{ where } \pi^* \!\!= \!\argmin_{\pi\in\Pi(\Z, \hZ)} \OT(\pi, \mathbf{D}(\Z,\hZ)),  
\end{equation}
where $\mathbf{D}$ denotes the pairwise distances between the columns in $\Z$ and $\hZ$, and $\OT$ is some suitable matching scheme. We use a general purpose optimal transport (IPOT~\cite{xie2020fast}) scheme to implement the alignment, which returns a permutation matrix $\pi^*$ that is used to align the matrices before comparing them using the $\ell_2$ distance.\footnote{We may also use a Hungarian matching scheme~\cite{jonker1986improving} to implement $\OT$ if the number of data instances is small, which is usually significantly faster than optimal transport methods. Note: our experiments suggest that a \emph{greedy} way to align is not useful---see Section~\ref{sec:ablation}.} We show this encoder control flow using solid red arrows in Fig.~\ref{fig:main}.

\noindent{\textbf{Intermediate Reconstruction:}}
To tackle difficulties (b) and (c) in the encoder design, which involve $E$ learning to invert the depth renderer, we use the output from the derenderer sub-module $\inv{\Gr}$ in $E$. Specifically, $\inv{\Gr}$ is forced to reconstruct the average-pooled feature map $\bF$ in~\eqref{eq:generator}. Let us denote this loss by $\loss^i_E = \enorm{\bF - \inv{\Gr}(\hvx)}^2$.

\noindent{\textbf{Pose Decoding:}} Although one could apply the above intermediate feature decoding strategy even to the pose decoder $\Gp$, it would not be very efficient to compare its output $\Lambda(\Gp(\hZ))$ to the rigid transforms produced during the generative process. This is because the geometric matrix that $\Lambda$ produces involves a rotation matrix, and thus optimizing would ideally require Riemannian optimization methods in the space of $\SO3$~\cite{absil2009optimization}, which is not well suited for standard optimization schemes such as Adam~\cite{kingma2014adam}. Furthermore, there may be several different geometric transformations that could achieve the same output~\cite{zhou2019continuity}. To avoid this technicality, we propose to learn the rigid transform indirectly, by avoiding exact reconstruction of the transform and instead asking it to have the desired outcome in the generative process. Specifically, we propose to use the $\hZ$ produced by the encoder, and use it as a noise matrix to produce a depth image $G(\hZ)$; this depth image is then compared with the depth image generated in the previous pass using $\Z$. The following loss, $\loss_E^p$, captures this idea:
\begin{equation}
    \loss_E^p = \lnorm{G(\Z) - G(E(\hvx))}.
\end{equation}
The above control flow is illustrated in Fig.~\ref{fig:main} by the dashed red arrows that go from noise vectors $\hvz$ through the pose decoder and over to the depth renderer, i.e., the output of $G$. 

\noindent{\textbf{Encoder Loss:}} We combine the above three losses when training the parameters of the encoder module (see the Supplementary Material for details on its architecture):
\begin{equation}
    \loss_E = \loss^a_E + \lambda_1\loss^i_E + \lambda_2\loss^p_E,
\end{equation}
where the $\lambda$'s provide weights to each type of loss.\footnote{We use $\lambda_1 = \lambda_2 =1$ in all our experiments.} When backpropagating the gradients on the encoder losses, we fix the generator parameters, as otherwise they will co-adapt with the encoder parameters, making training unstable.

\noindent{\textbf{Learning the Implicit Object Template:}} The template is implemented as a weight tensor, learned via backpropagation gradients from the above loss. During training, backpropagation reverses all of the arrows in Fig.~\ref{fig:main}.

\subsection{\name Inference}
\vspace{-6pt}
\label{sec:inference}
At inference time, we assume to be given only a depth image consisting of multiple instances of the rigid object; our goal is to segment the instances and render each instance separately, while producing an instance segmentation on the input. To this end, our inference pipeline resembles the generative process, but with some important differences as illustrated in Fig.~\ref{fig:inference}. Specifically, for inference we input the multiple-instance depth image to the instance pose encoder module $E$, which produces a set of latent vectors $\hZ$. Each $\hvz\in\hZ$ is input individually into the trained single-instance generator $\Gs$, the output of which is rendered using $\Gr$ to form a single-instance depth image that corresponds to $\hvz$. We emphasize that in the inference phase, the depth image renderer sits within the single-instance generation phase---this contrasts with the
training setting, in which the renderer takes as input the aggregated feature tensor $\bF$.  Once the single instances are rendered, as shown in Fig.~\ref{fig:inference}, we use a depth-wise max pooling on these instance depth images for inter-instance occlusion reasoning, followed by thresholding (and applying basic image filters to) the single instances. Thresholding removes any bias introduced during depth rendering. To produce the pixel-wise segmentation, we use the index of the generated instance that is selected for a given pixel. 

\subsection{Training Pipeline}
\vspace{-6pt}
We train our full framework, including the \name generator $G$, discriminator $D$, and Encoder $E$, by minimizing the sum of all the losses given by:
\begin{equation}
    \loss = \loss_D + \loss_E + \loss_G.
\end{equation}
The gradients for the various modules are computed using PyTorch autograd. We use Adam for training all our models, with a learning rate of 0.0002, $\beta_1=0.5$, and $\beta_2=0.99$. 
\section{Experiments and Results}
\vspace{-6pt}
\label{sec:expts}
In this section, we present experiments demonstrating the empirical benefits of \name on the task of instance segmentation. We will first introduce our new synthetic dataset, \emph{Insta-10}, on which most of our experiments are based. We then introduce a real-world dataset that we collected to evaluate the application of our method  on (naturally noisy) depth images of real objects.

\begin{table*}[h]
    \centering
    \begin{tabular}{c|c|c|c|c|c|c|c|c|c|c|c}
    Method & Nut & Stop. & Cyl. & Bolt & Cone & Conn. & 5-pin & Obj01 & Obj14 & Obj05 & Avg mIoU\\
    \hline
     \multicolumn{12}{c}{Non-Deep Learning Methods}\\
    \hline
    K-Means & 0.64 & 0.297 & 0.7 & 0.18 & 0.35 & 0.554 & 0.628 & 0.208 & 0.496 & 0.59 & 0.464\\
    Spectral Clustering~\cite{ng2002spectral} & 0.56 & 0.36 & 0.54 & 0.22 & 0.41 & 0.56 & 0.58 & 0.25 & 0.47 & 0.57 & 0.452 \\
    GrabCut~\cite{rother2004grabcut}+KMeans& 0.572 & 0.232 & 0.572 & 0.472 & 0.231 & 0.519 & 0.497 & 0.597 & 0.557 & \textbf{0.605} & 0.486\\
    GraphCut~\cite{boykov2004experimental} & 0.569 &	0.1	 & 0.589 & 	0.447	 & 0.12	 & 0.476 &	0.12	& 0.597	& 0.540	 & 0.511 & 0.373\\
    \hline
     \multicolumn{12}{c}{Deep Learning Methods}\\
    \hline
    Wu et al.~\cite{wu2019unsupervised}& 0.45 & 0.28 & 0.57 & 0.27 & 0.33 & 0.38 & 0.43 &  0.23 & 0.44 & 0.57 & 0.385\\
    IODINE~\cite{greff2019multi} & 0.026 & 0.059 & 0.019 & 0.040 & 0.089 & 0.032 & 0.034 & 0.058 & 0.053 & 0.118 & 0.053 \\
    Slot Attn.~\cite{locatello2020object} & 0.375 &	0.276 & 	0.535 &	0.43 &	\textbf{0.68} &	\textbf{0.662}	& 0.628	 & 0.655 & \textbf{0.622} & 0.481 &0.535\\
    \hline
    \name (2D) (ours) & 0.215 & \textbf{0.365} & 0.258 & 0.524 & 0.435 & 0.585 & 0.628 & 0.365 & 0.286 & 0.532  & 0.419\\
    \name (3D) (ours)    & \textbf{0.773} & 0.301 & \textbf{0.760} & \textbf{0.539} & 0.47 & 0.655 & \textbf{0.642} & \textbf{0.686} & 0.591 & 0.483  & \textbf{0.590}
    \end{tabular}
    \caption{Mean IoU (mIoU) between the segmentation masks predicted by each method and the ground-truth masks.}
    \vspace{-8pt}
    \label{tab:soa}
\end{table*}

\begin{table}[ht]
    \centering
    \begin{tabular}{c|c|c}
    Generator Loss & Bolt & Obj01 \\
    \hline
        $\loss^a_E$ (OT) + $\loss^i_E + \loss^p_E$ & \textbf{0.424} & 
        \textbf{0.686} \\
        $\loss^a_E$ (greedy) + $+\loss^i_E + \loss^p_E$ &0.383 & 0.664\\
        $\loss^a_E$ (OT) + $\loss^i_E$  & 0.312& 0.360\\
        $\loss^a_E$ (OT) & 0.303& 0.402\\
    \end{tabular}
    \caption{Ablation study on the various losses used in \name generator and the mIoU achieved on two classes.}
    \vspace{-8pt}
    \label{tab:ablative_losses}
\end{table}

\begin{table}[ht]
    \centering
    \begin{tabular}{c|c}
    Method & mIoU  \\
    \hline
        KMeans & 0.797\\
        Spectral Clustering & 0.668\\
        Graph Segmentation~\cite{felzenszwalb2004efficient} & 0.436\\
    \hline
        \name & \textbf{0.857}
    \end{tabular}
    \caption{Results on real-world data collected using a robot.}
    \vspace{-12pt}
    \label{tab:realdata}
\end{table}
\begin{figure}
    \centering
     \subfigure[Datasize]{\label{fig:inc_datasize}\includegraphics[width=4cm,trim={0.6cm 0.3cm 1cm 1cm},clip]{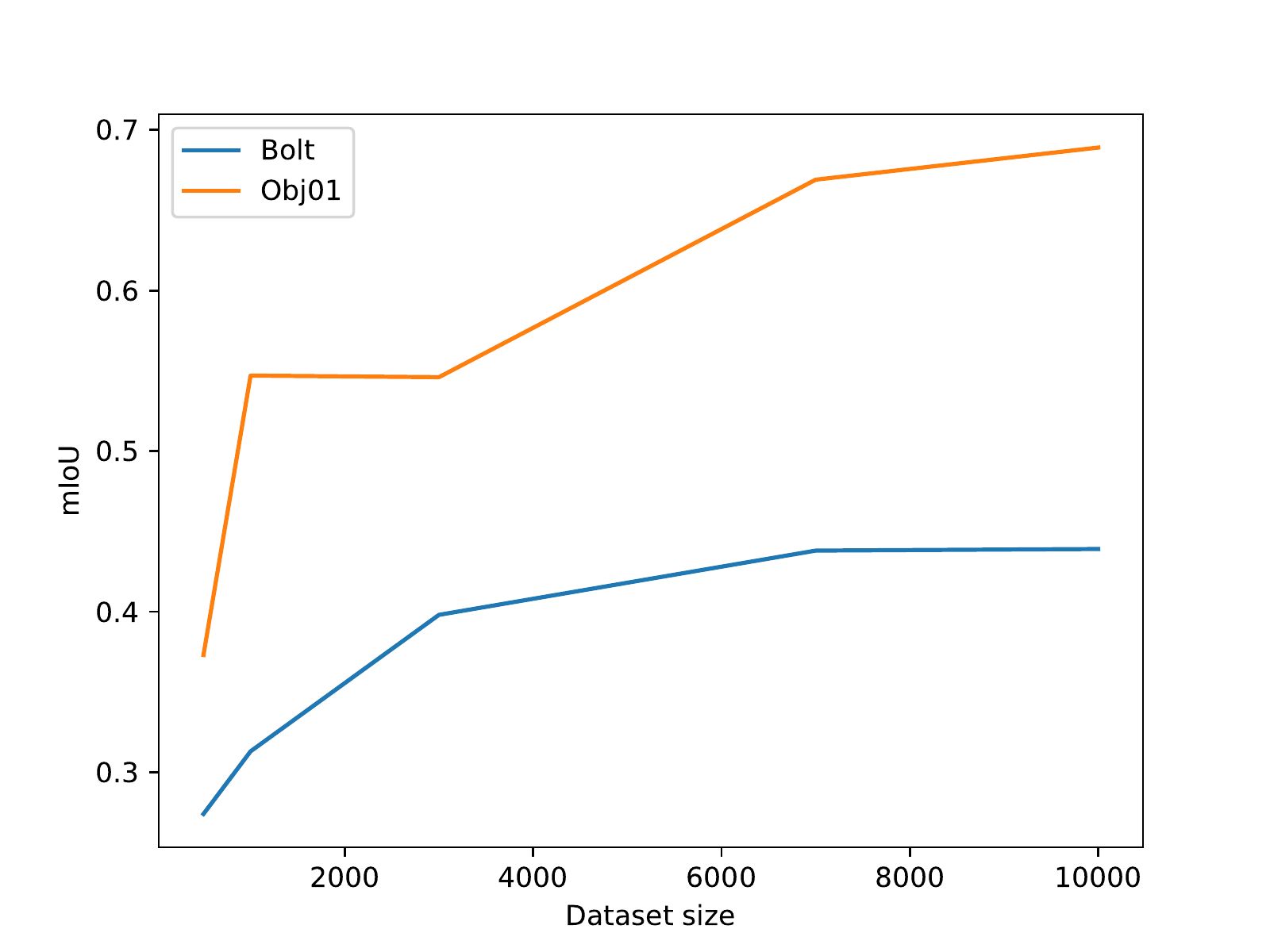}}
    \subfigure[Instances]{\label{fig:inc_instances}\includegraphics[width=4cm,trim={0.6cm 0.3cm 1cm 1cm},clip]{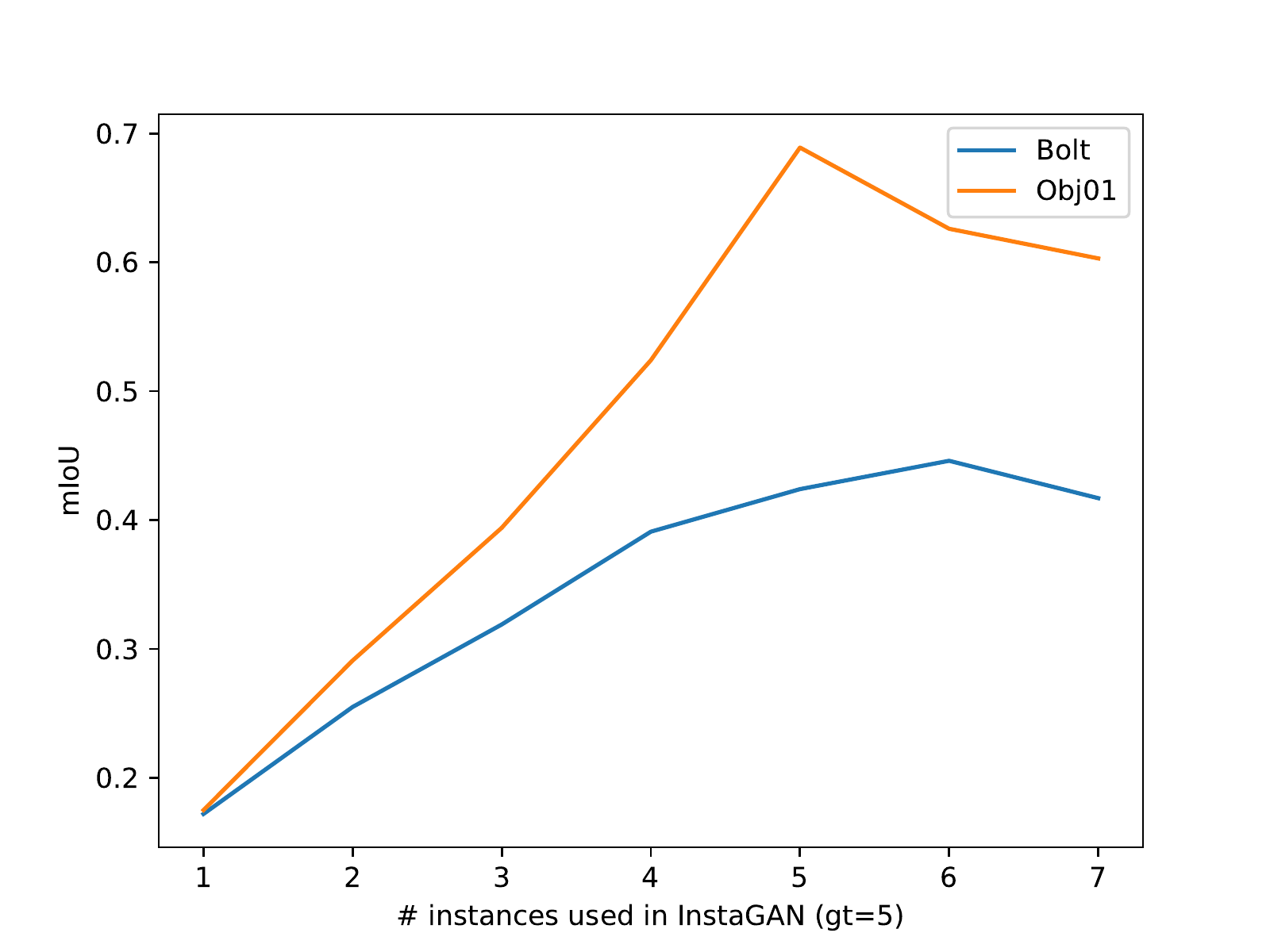}}
    \caption{mIoU versus (a) training dataset size, (b) number of instances $n$ in model (ground truth has 5 instances).}
    \vspace{-12pt}
\end{figure}

\noindent\textbf{Insta-10 Dataset:}
While there are several real-world datasets used for instance segmentation, such as MSCOCO~\cite{lin2014microsoft}, and CityScapes~\cite{cordts2016cityscapes}, they typically involve background objects, and other \emph{stuff} that are unrelated to the objects to be segmented. In addition, datasets such as CLEVR~\cite{johnson2017clevr} are proposed for visual reasoning tasks, and thus may not fully analyze the segmentation quality. 
To fill this gap, we introduce Insta-10, a large-scale dataset consisting of depth images of multiple instances of a CAD object model. We remove color and texture from the instances, to analyze the segmentation performance under the difficult condition in which there are minimal attributes other than shape. This is inspired by the observation that most industrial objects do not usually have textures~\cite{hodan2017t}, in addition to the intuition that sometimes RGB could distract a shape-based segmenter.

To create the dataset, we used 10 CAD object models (3 from the T-less dataset~\cite{hodan2017t} and 7 from our own library). We use the PhysX physics simulator\footnote{\url{https://developer.nvidia.com/physx-sdk}} to simulate sequentially dropping objects into a bin, producing synthetic multiple-instance depth images. We used 5 instances of the same object in each depth image, yielding substantial inter-instance occlusion, and we selected the bin width so that instance segmentation was challenging but not too hard (even for humans). In addition to the depth images, we provide the point clouds associated with each image and ground truth instance segmentation masks; these masks are used for only evaluation, not training. We collected 10K images per object, for a total of 100K depth images in the entire dataset, each with dimension $224\times 224$. Sample images are in Figs.~\ref{fig:qual_results} and~\ref{fig:qual_soa}.

\noindent\textbf{Real-World Depth Images Using a Robot:}
Apart from the synthetic Insta-10 dataset, we also analyze the adaptability of our scheme to practical settings. For this experiment, we used a box containing  4 identical wooden blocks (see Fig.~\ref{fig:realdata}), of which depth images were captured using an Intel RealSense Depth Camera (D435). To produce multiple diverse images consisting of varied configurations of the blocks, we programmed a Fetch robot~\cite{wise2016fetch} to shake the box between images. We collected 3,000 depth images using this setup, of which we hand-annotated 62 images that we reserved for evaluation. The depth images from this setting are very noisy, and as a result, often the shapes of the objects do not appear to be identical.

\begin{figure}
    \centering
    \includegraphics[width=8cm,trim={0.cm 12.9cm 0cm 1cm},clip]{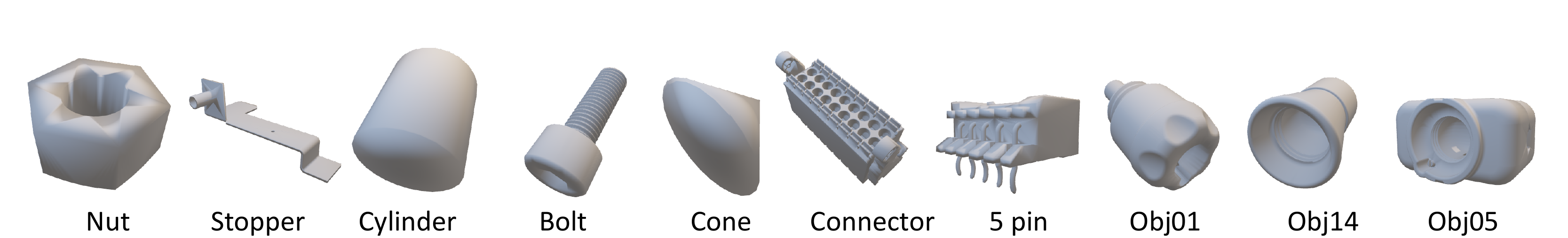}
    \includegraphics[width=8cm,trim={2cm 12.75cm 6cm 3cm},clip]{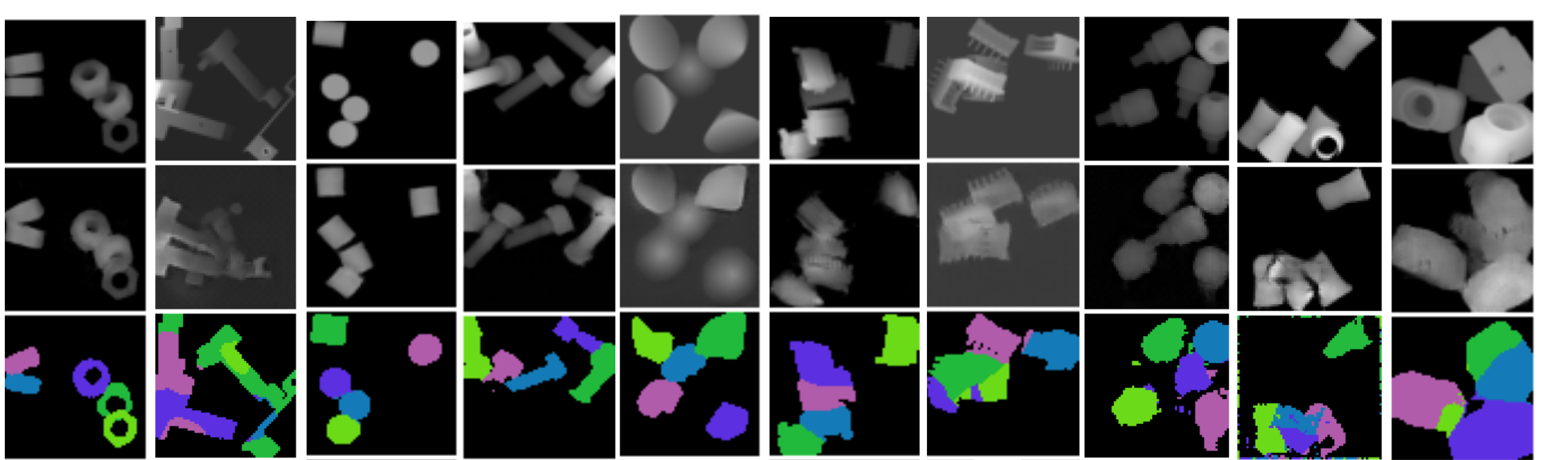}
    \caption{Qualitative results on Insta-10 objects. First row: CAD models used to produce Insta-10. Second row: the input depth images. Third row: rendered depth image by \name. Fourth row: the predicted segmentations by \name.}
    \vspace{-6pt}
    \label{fig:qual_results}
\end{figure}
\begin{figure}
    \centering
    \includegraphics[width=8cm,trim={1cm 8.7cm 1cm 8cm},clip]{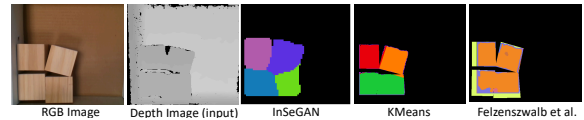}
    \caption{Qualitative results on real data. We show the RGB image, the noisy depth input, and the segmentations produced.}
    \vspace{-6pt}
    \label{fig:realdata}
\end{figure}
\begin{figure}
    \includegraphics[width=8cm,trim={3cm 10.3cm 9cm 4.5cm},clip]{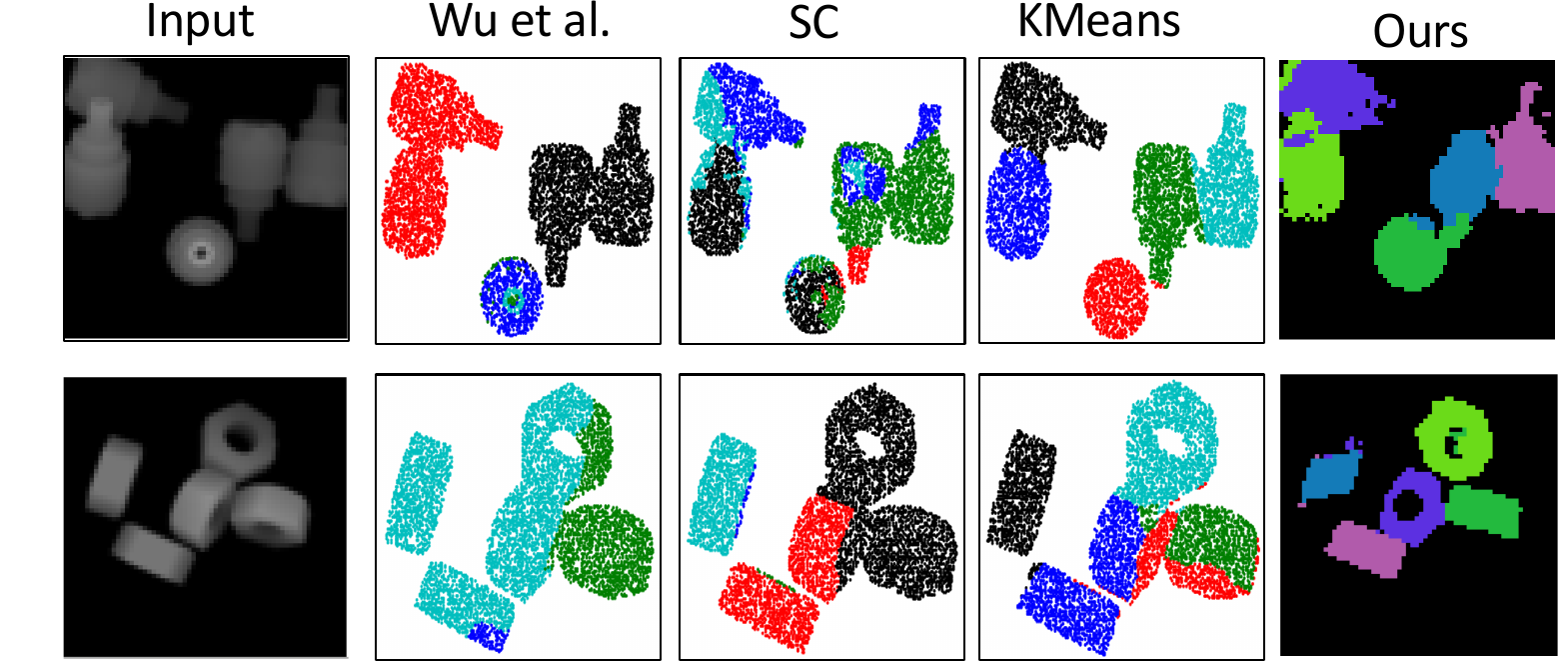}
    \caption{Qualitative comparisons against other methods.}
    \vspace{-12pt}
    \label{fig:qual_soa}
\end{figure}

\noindent\textbf{Evaluation Metric and Experimental Setting:}
To evaluate our scheme, we use mean intersection-over-union (mIoU), a standard metric for semantic segmentation. For training and evaluation, we split the data subsets associated with each class into a training, validation, and test set. In the Insta-10 dataset, we use 100 randomly selected images for validation in each class. For the test set, 
we selected 100 images on which KMeans fails, thereby avoiding segmentations that are trivial for standard methods. 

\noindent\textbf{Performance Analysis:} On the Insta-10 dataset, we compare our method on both non-deep and deep learning methods. The non-deep methods include classic segmentation algorithms~\cite{ng2002spectral,boykov2004experimental,rother2004grabcut}.  The deep learning comparisons include: (i) Wu et al.~\cite{wu2019unsupervised}, which is most similar to ours; (ii) IODINE~\cite{greff2019multi}, which was proposed for scene decomposition rather than instance segmentation; and (iii) Slot Attention~\cite{locatello2020object}. We use the public code for (ii) and (iii), using their default hyper-parameters. In Table~\ref{tab:soa}, we show these results. We find that for most object classes (6/10), \name outperforms all other methods.

On the Stopper class, which is the most difficult, \name outperforms all other methods except for spectral clustering. Overall, \name demonstrates a relative improvement of 9.3\% over the best-performing previous method (averaged across all 10 classes). We found that the recent method of IODINE~\cite{greff2019multi} fails on our images, perhaps because it is designed for scene decomposition tasks. From Table~\ref{tab:realdata}, we see that our method generalizes to real data as well. In Fig.~\ref{fig:qual_results}, we show several qualitative results produced by \name.  More results are provided in the Supplementary Material.

\subsection{Ablation Studies}
\label{sec:ablation}
\vspace{-6pt}
In this section, we analyze each component in our design, empirically justifying its importance.

\noindent\textbf{Is the 3D Generator Important?}
To answer this question, we replace the 3D modules in \name (3D implicit template, pose encoder, and STN) by 2D convolutions and upsampling layers, similar to those used in the encoder and discriminator. In Table~\ref{tab:soa}, we provide comparisons of the 3D and 2D GANs on the Insta-10 dataset. Results show that our 3D generator is significantly better than a 2D generator. 

\noindent\textbf{Are all the losses important?}
There are three losses in the \name generator: (i) $\loss^a_E$, the alignment loss, (ii) $\loss^i_E$ on the intermediate feature maps, and (iii) $\loss^p_E$ between the generated depth image and the re-generated depth image. For (i), we compare a \emph{greedy} choice for alignment vs. using optimal transport. We provide ablative studies on two object classes, Bolt and Obj01. As is clear from Table~\ref{tab:ablative_losses}, using a greedy alignment leads to lower performance. Further, we find that using $\loss^p_E$ is empirically very important, yielding 10--20\% performance improvement. Our analysis confirms the importance of all of the losses used in our architecture.

\noindent\textbf{Do we need all training samples?}
In Fig.~\ref{fig:inc_datasize}, we plot the performance versus increasing the number of data samples; i.e., we train on a random subset of the 10K depth images in the training set. Clearly more training data is useful, but this increment appears to be dependent on the object class. 

\noindent\textbf{Wrong number of instances?}
 In Fig.~\ref{fig:inc_instances}, we plot the performance versus increasing the number of instances used in \name; i.e., we increase $n$ from 1 to 7. This is a mismatch from the true number of instances (5 in every depth image). The plot shows that \name performs reasonably well when the number of instances is close to the ground truth. In the Supplementary Material, we show how to handle an unknown number of instances $n$ in each image. 
 
\section{Conclusion}
\vspace{-6pt}
\label{sec:conclude}
In this paper, we presented \name, a novel 3D GAN to solve unsupervised instance segmentation. We find that by pairing the discriminator with a carefully designed generator, the model can reconstruct individual object instances even under clutter and severe occlusions.  We introduce a new large-scale dataset, which we are making publicly available, to empirically analyze our approach. Our method demonstrates state-of-the-art results, generalizing well to real-world images. 

{\small
\bibliographystyle{ieee_fullname}
\bibliography{instagan}
}
\pagebreak
\appendix

\section{\name: Insights and Why it Works}
A curious reader of our work might ask, \emph{How does the network learn to disentangle the depth image into each instance poses and the implicit template? In particular, how does it learn to disentangle the pose of each instance from the depth image into a separate latent vector in $\hZ$? And, why does the network learn an implicit template model that represents a single instance, rather than multiple instances within a single template?}
This is, we believe, because of the way the generator-discriminator pipeline is trained. For example, let us assume for a moment that a single latent noise vector $\vz$ controls more than one (or in the extreme, all) of the instances in a depth image. As $\vz$ is randomly sampled from a distribution, it is unlikely that only some of the vectors in $\Z$ (the collection of $n$ $\vz$ vectors used as input to the generator) would render the instances and some would not, given that aggregation of all the generated instances should match up to the number of instances in the input data---a requirement that the discriminator will eventually learn to verify in the generated images. Further, given that the object appearances are varied, it is perhaps easier for the generator to learn to render the appearance of a single instance than to capture the joint appearance distribution for all instances, which could be very large and diverse. 

\begin{figure}[t]
    \centering
    \includegraphics[width=8cm,trim={5cm 3cm 5.4cm 3cm},clip]{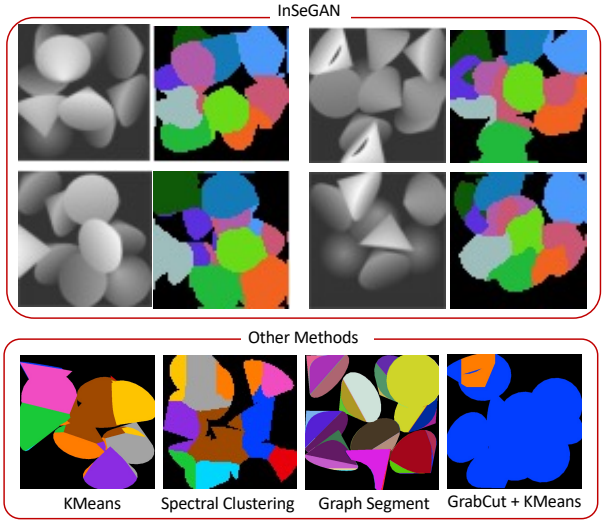}
    \caption{Qualitative results of \name segmentation on the synthetic 10-cones dataset. We also show example segmentations produced by other methods.}
    \label{fig:cones10}
    \vspace*{-10pt}
\end{figure}

\section{How is the Implicit Template Learned?}
Note that the template is learned jointly with the rest of the modules. The teplate is implemented as a PyTorch weight tensor and is updated with the backpropagation gradients from the losses. Simply put, when training the setup, all the arrows in Figure 2 gets reversed, thus training the template along with all of the other weights in the network.

\section{Using Arbitrary Number of Instances $n$?}
\begin{figure}[h]
\centering
\includegraphics[width=8cm,trim={1cm 1cm 18cm 1cm},clip]{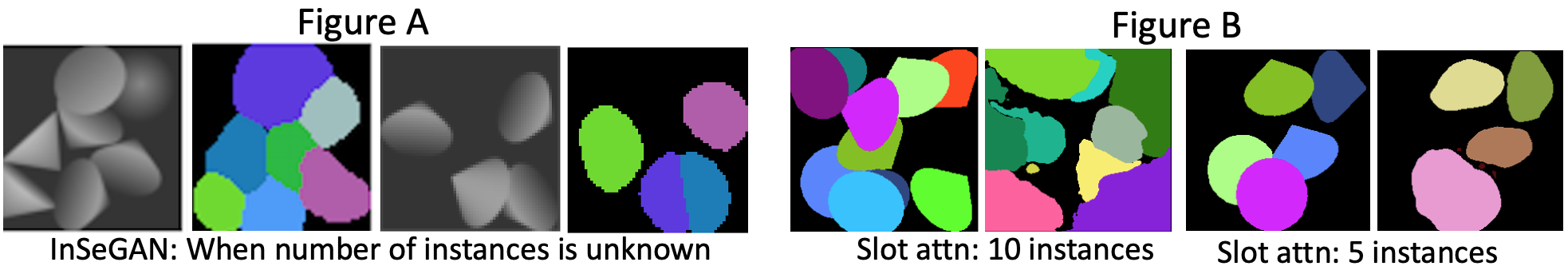}\\
\caption{InSeGAN results when we use the same model to learn distributions of images with varying number of instnaces. The results show segmentation visualizations when we used 4--9 instances in each of the depth images.}
\label{fig:n-varying}
\end{figure}

It is straightforward to extend InSeGAN for an arbitrary number of instances $n$, which we do by using training images with varying numbers of instances. Assuming a max of $n$ instances in the training images, we have InSeGAN sample a random number ($\leq n$) of pose vectors at the input in Fig. 2 (paper). Further, we also add a simple module that predicts the number of instances in the rendered image, which is used to produce that many pose vectors. A loss is enforced that ensures the number of sampled pose vectors and the number of estimated pose vectors (by the instance pose encoder) are the same. The rest of the pipeline stays the same. At test time, the input depth image is passed through the instance pose encoder, alongside the number of estimated instances (by the additional module), and each of the produced instance poses are decoded individually to produce the segmentations. We implemented this variant of our scheme and found that the GAN successfully learns to match the new distribution, which is that of depth images with varied instance count and produces instance segmentations for arbitrary number of object instances. In Fig.~\ref{fig:n-varying}, we show results on the Cone class when we vary the count between 4 and~9. On these data, we achieved 45.1\% mIoU.

\section{Synthetic Setting with 10 Cones}
As introduced in Figure 1 of the main paper, we also explored the scalability of \name to depth images with more than 5 instances. Similarly to how we produced the synthetic Insta-10 dataset with $n=5$ instances in each category, we produced an additional dataset using $n=10$ instances of cones, to explore how well our model handles the more difficult case of depth images with twice as many instances. As in the Insta-10 dataset, all 10 instances were randomly dropped into a bin in sequence using a physics simulator. Similar to each category in Insta-10, we created 10,000 depth images with 10 cones each, of which we used 100 for validation and 100 for testing. We did not use K-Means to select difficult examples for our test set in the 10-instance setting, because the increased number of cones means that every depth image in the set is cluttered and quite challenging. We trained our \name model with exactly the same setting and hyperparameters (except for the number of instances $n$). Qualitative results are provided in Figure~\ref{fig:cones10}. In Table~\ref{tab:cones10}, we quantitatively compare the performance of \name on this dataset.

\begin{table}[h]
    \centering
    \begin{tabular}{c|c}
    Method & mIoU  \\
    \hline
        KMeans & 0.302\\
        Spectral Clustering & 0.324\\
        Superpixels~\cite{wei2018superpixel} & 0.398\\
        GrabCut+KMeans & 0.021\\
    \hline
        \name & \textbf{0.501}
    \end{tabular}
    \caption{Numerical comparison of InSeGAN vs. other methods on challenging dataset with $n=10$ cones in each image.}
    \vspace{-12pt}
    \label{tab:cones10}
\end{table}

\section{Qualitative Results on Real Data}
\begin{figure}
    \centering
    \includegraphics[width=8cm,trim={1.5cm 5cm 5cm 3cm},clip]{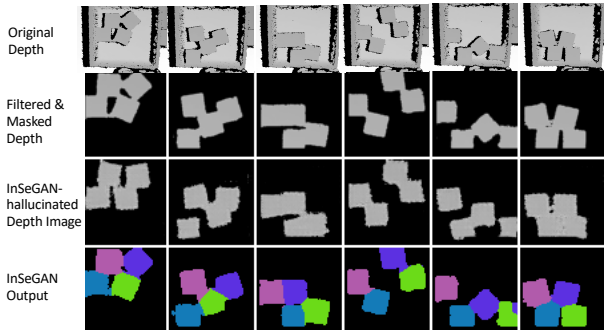}
    \caption{Qualitative results on instance segmentation of real data. We show the original depth images collected using a robot (first row), the output of a masking and filtering step we do to clean up the inputs (second row), the depth images hallucinated/rendered by \name (third row), and the segmentations (fourth row).}
    \label{fig:realdata_more_quals}
\end{figure}

In Figure~\ref{fig:realdata_more_quals}, we show qualitative results of instance segmentation on real data. We also highlight the preprocessing steps we follow to apply \name to this dataset, which are necessary because the input depth images are very noisy (e.g., jitter/spurious noise in the depth sensor, and hole-filling that the sensor algorithm implicitly applies). These steps often alter the object shape, for example, merging two adjacent instances to appear as a single large object. To use these depth images in our setup, we first masked out the surrounding region (everything outside of the bin). This is possible because the bin is always at the same location. After applying the mask, we thresholded the z values in the depth image to only show z values greater than half the height of a block instances. This provided a relatively clean depth image, reducing the jitter and other artifacts (see Figure~\ref{fig:realdata_more_quals}). Next, we applied \name to these preprocessed images. The qualitative and quantitative results (in the main paper) show that \name is very successful in segmentation on these images (~85\% mIoU). 

\section{Data Collection Setup}
In this section, we detail our synthetic and real-world data collection setups.

\subsection{Physics Simulator and Depth Image Generation}
As described in the main paper, we use the NVIDIA PhysX physics simulator\footnote{\url{https://developer.nvidia.com/physx-sdk}} to create our Insta-10 dataset. A screenshot of this simulator software setup is shown in Figure~\ref{fig:simulator}. Specifically, the simulation consists of a virtual bin of a suitable size and depth (depending on the size of the object) into which virtual instances (Cones in the figure, for example) are dropped sequentially from random locations above the bin. Next, an overhead simulator depth camera captures the depth image associated with the instances. A snapshot of the instance segmentation of the five objects is shown in the figure (right). The simulator automatically takes care of avoiding intersecting objects (because it is physically impossible) and accounts for occlusions. It takes approximately 2 seconds to generate one depth image using this setup with 5 identical object instances.

\subsection{Robotic Collection of Real-World Depth Images}
We first describe our robotic experiment system, then explain how we use it to collect more than 3,000 real-world depth images for testing our approach. Our experiments are carried out on a Fetch robot~\cite{wise2016fetch} equipped with a 7-degree-of-freedom (7-DOF) arm, and we use ROS~\cite{quigley2009ros} as our development system. The Fetch robotic arm is equipped with a stock two-fingered parallel gripper. Mounted above the box is a downward-pointing Intel RealSense Depth Camera (D435), which consists of depth sensors, RGB sensor, and infrared projector. The camera, which is attached to a Noga magnetic base, provides a depth stream output with resolution up to 1280 $\times$ 720 resolution  of the scene with which the Fetch robot interacts. For trajectory planning, we use the Expansive Space Tree (EST) planner~\cite{phillips2004guided}. Note that during the experiments, human involvement is limited to switching on the robot, configuring the planner, and placing the objects in the box arbitrarily. Apart from this initialization, our robotic pipeline has no human involvement in the process of data collection. 

\begin{figure}[t]
\centering
\includegraphics[scale=0.3]{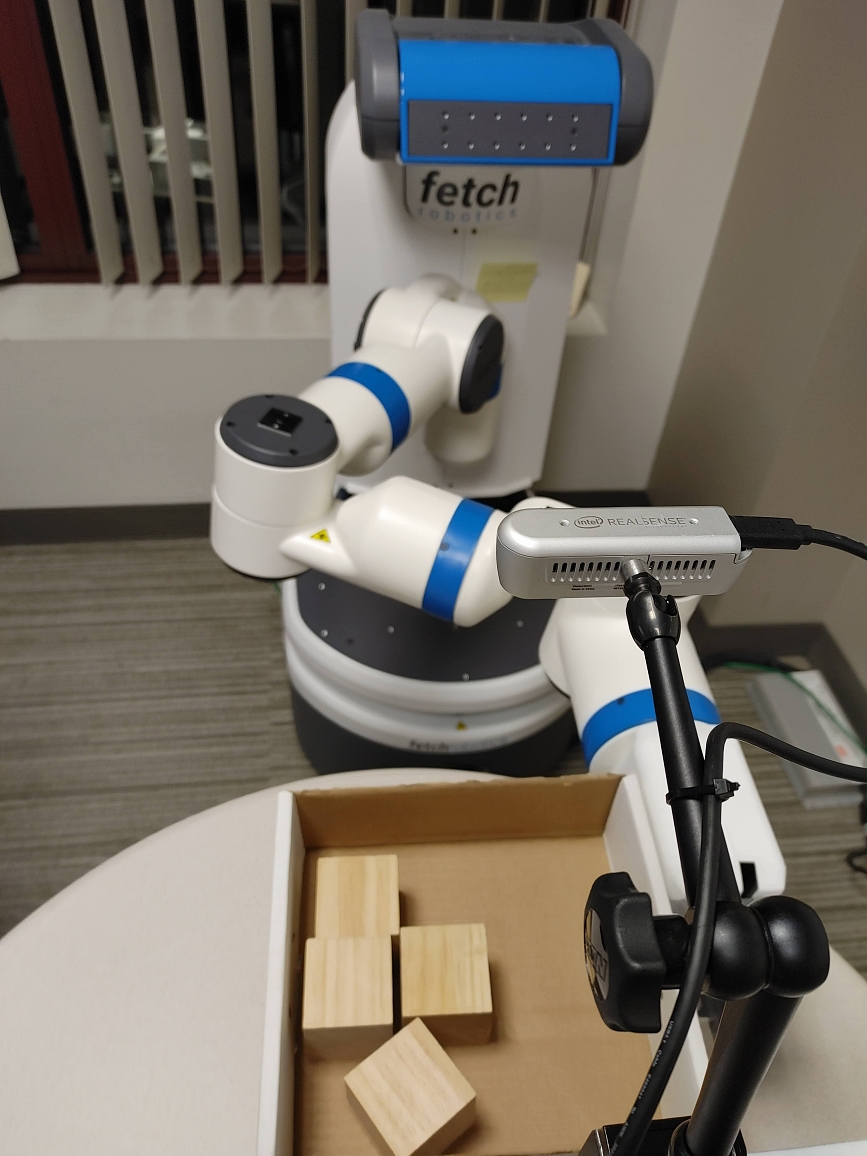}
\caption{Robotic data collection system used to acquire real-world depth images.}
\label{fig:robot}
\end{figure}

\begin{figure}[t]
\centering
\includegraphics[width=4cm,trim={0cm 0cm 8cm 0cm},clip]{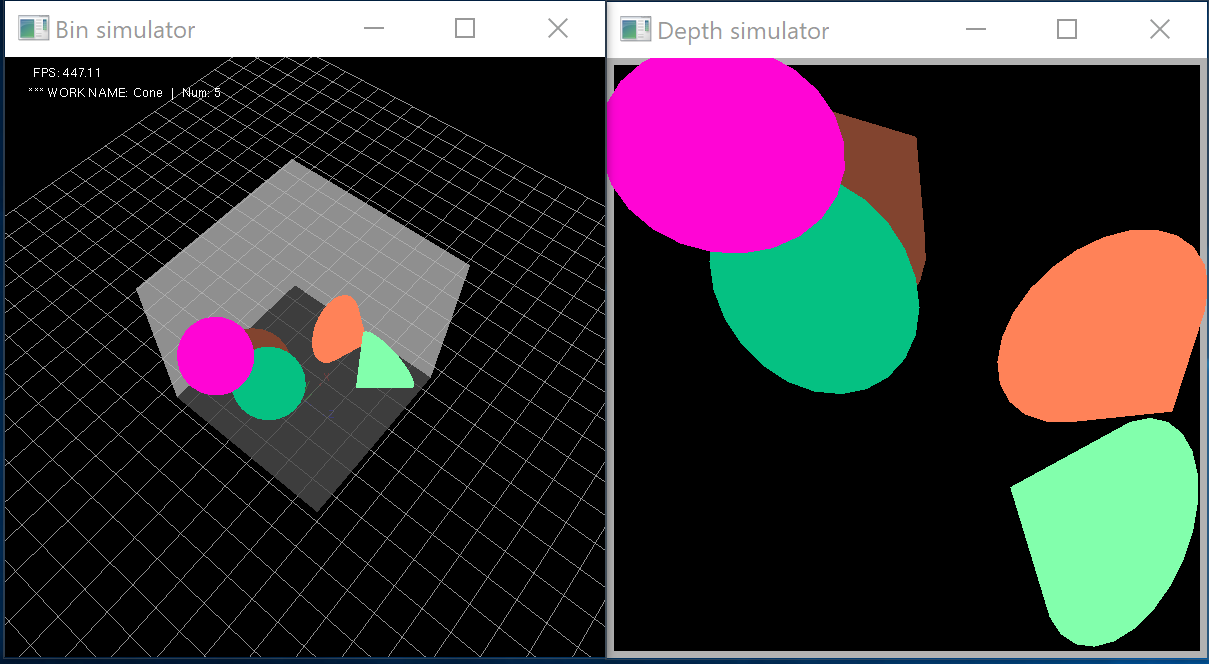}
\includegraphics[width=4cm,trim={8cm 0.1cm 0cm 0cm},clip]{./figs/simulator.png}
\caption{An illustration of the physics simulator that we use to render our synthetic dataset, Insta-10. \emph{Left:} the simulated bin into which the identical objects (e.g., Cone) are dropped. \emph{Right:} The ground-truth instance segmentation masks for each of the instances. We use the depth images associated with these instances for training \name, so that at inference time, segmentation masks are recovered. The ground-truth instance segmentation masks are not used for training---they are only used for testing our unsupervised method.}
\label{fig:simulator}
\end{figure}

The data collection setup is depicted in Fig.~\ref{fig:robot}. The workspace is first set up with a box with plain background, and the box is fitted with a handle that is grasped by the Fetch robotic arm. Four identical wooden blocks are placed inside the box in random pose configurations. A single instance of a trial proceeds as follows: A depth image of the box is captured by the depth camera and recorded to a disk. The robot then initiates the trajectory planner, and the robotic arm executes a motion trajectory to tilt-shake the box randomly in the clockwise or anticlockwise direction such that the motion is collision free, then returns the box to its original location. The degree of tilt shake is also randomized (up to a specified maximum to prevent the blocks falling out of box). Multiple trials are executed in succession for the robot to autonomously record the dataset, with four cycles per minute. 

\section{Network Architectures}
In this section, we will detail the neural architectures of the three modules in \name: (i) the Encoder, (ii) the Discriminator, and (iii) the Generator. 

\begin{figure*}[ht]
\centering
\includegraphics[width=16cm,trim={0cm 7cm 0.2cm 4cm},clip]{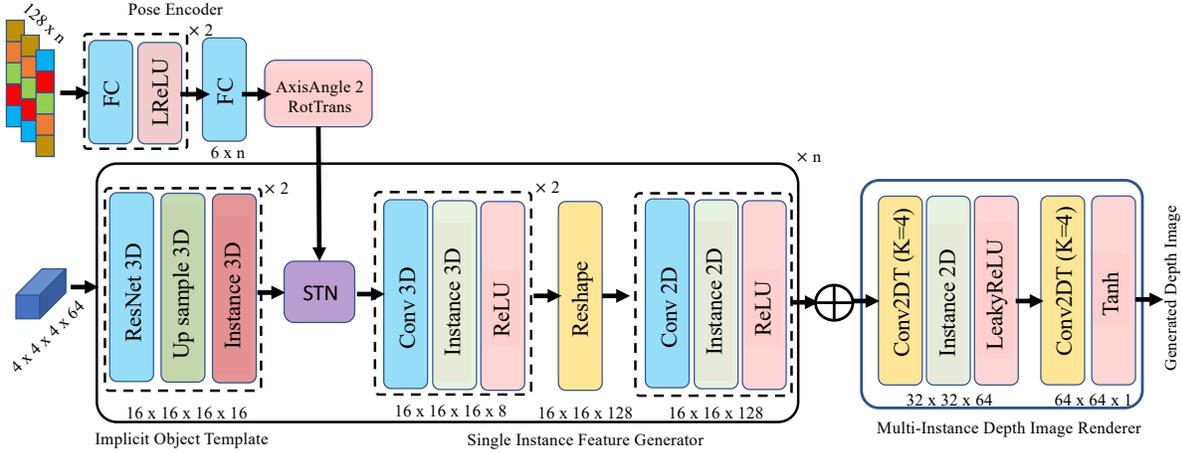}
\caption{Detailed architecture of \name generator.}
\label{fig:generator}
\end{figure*}

\noindent{\textbf{Generator:}}
In Fig.~\ref{fig:generator}, we provide the detailed architecture of our \name Generator. It has five submodules: (i) A pose decoder, which takes $n$ random noise vectors $\vz_i\in\reals{128}\sim \normal(0,\eye_{128})$, where $n=5$ in our setup, and produces 6-D vectors that are assumed to be axis-angle representations of rotations and translations~\cite{zhou2019continuity} (three dimensions for rotation and three for translation). Each 6-D vector is then transformed into a rotation matrix and a translation vectors, to produce an element in the special Euclidean group $\bigl( \SE3 \bigr)$. (ii) A 3D implicit template generation module, which takes a $4\times 4\times 4 \times 64$ dimensional tensor (representing an implicit 3D template of the object) as input, then up-samples in 3D using ResNet blocks and 3D instance normalization layers to produce a $16 \times 16 \times 16 \times 16$ feature maps. (iii) A spatial transformer network (STN)~\cite{jaderberg2015spatial}, which takes as input the 3D implicit template and the geometric transform for every instance, then transforms the template, resamples it, and produces a transformed feature map of the same size as its input. (iv) A single-instance feature generator module, which reshapes the transformed template feature and produces single-instance 2D feature maps (each of size $16 \times 16 \times 128$). (v) A depth renderer module that takes an average pool over the $n$ feature maps representing the $n$ instances, and renders a multiple-instance depth image from the pooled feature map. 


The 3D implicit template loosely follows the architecture of a HoloGAN~\cite{nguyen2019hologan}, but differs in that we do not use any stochastic modules (via MLP) that were critical in their framework to produce stochastic components in the generated images (RGB images, in their case). We found that using noise vectors as in HoloGAN failed in our setup, causing us to lose the ability to disentangle instances. 

\begin{figure}[ht]
    \centering
    \subfigure[Encoder]{\label{fig:encoder}\includegraphics[width=8cm,trim={1cm 8.5cm 6cm 7.6cm},clip]{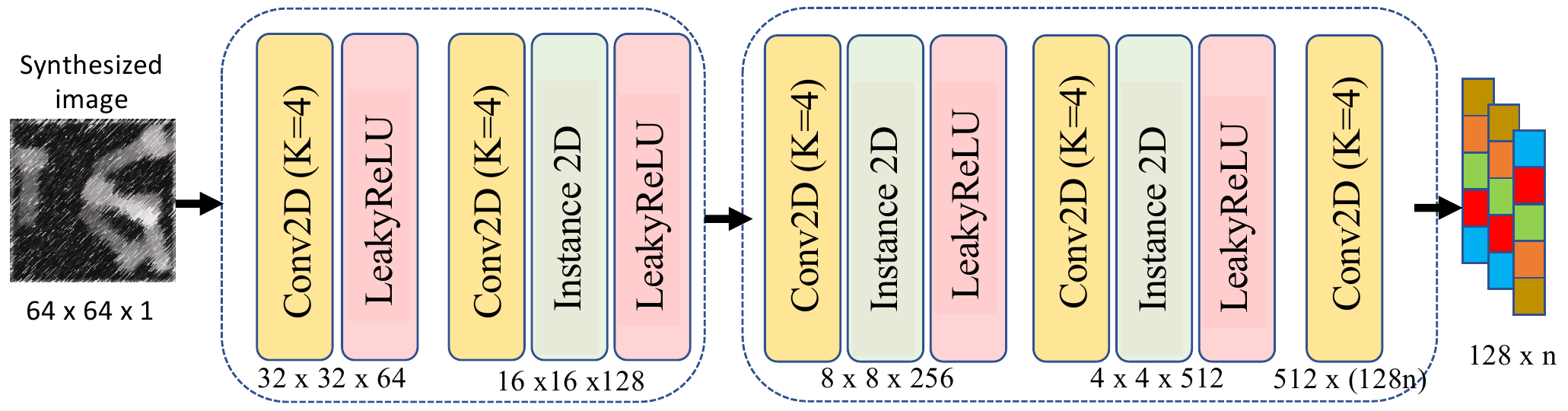}}
    \subfigure[Discriminator]{\label{fig:discriminator}\includegraphics[width=8cm,trim={1cm 8.6cm 6.5cm 8.0cm},clip]{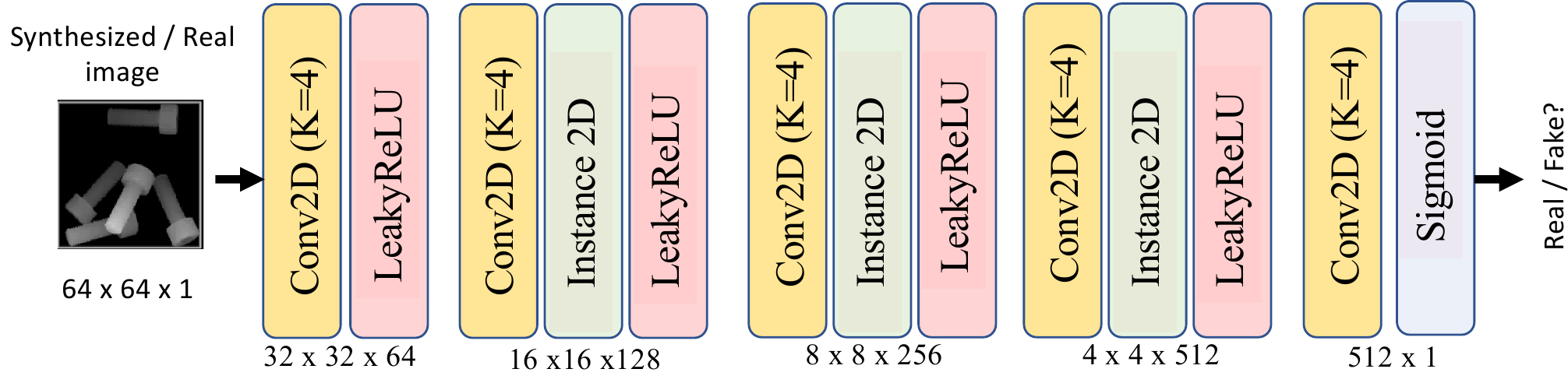}}
    \caption{(a) depicts detailed architecture of our Encoder module, and (b) shows our Discriminator module.}
    \label{fig:encoder_and_discriminator}
\end{figure}
\noindent{\textbf{Encoder and Discriminator:}}
In Fig.~\ref{fig:encoder_and_discriminator}, we show the neural network used in our Encoder and our Discriminator. They loosely follow similar architectures, except that the Discriminator takes a $64\times 64$ depth image (either generated or from the real examples) as input and produces a scalar score, while the encoder takes a generated depth image and produces the $n$ pose instance vectors as output. We use 128-D noise vectors when generating the images, and thus the Encoder is expected to produce 128-D features as output (one 128-D feature for each instance). Both the Discriminator and the Encoder use 2D convolutions, leaky ReLU activations, and 2D instance normalization~\cite{ulyanov2016instance} modules.

\subsection{Implementation Details and Training Setup}
Our \name modules are implemented in PyTorch. As alluded to above, we generate $224\times 224$ depth images using our simulator; however, we use $64\times 64$ images in our \name pipeline. To this end, each $224\times 224$ image is rescaled to $64\times 64$ and normalized using mean subtraction and normalization by the variance. For training, we use horizontal and vertical image flips for data augmentations. We do not use any other augmentation scheme. 

\subsection{Evaluation Details}
For our evaluations, we use the mean IoU (mIoU) metric between the ground truth instance segments and the predicted segmentations. Specifically, for each ground truth segment, we find the predicted segment that is most overlapping with this segment, and compute their intersection-over-union (IoU); we then use every segment's IoU to compute the mean IoU over all segments.

\noindent{\textbf{Training:}} We train our modules for 1000 epochs using a single GPU; each epoch takes approximately 30 seconds on the $\sim$10,000 training samples for each object. We use the Adam optimizer, with a learning rate of $2\times 10^{-4}$, and $\beta_1=0.5$. We use 128-D noise samples from a Normal distribution for the noise vectors, and a batch size of 128 samples. 

\begin{figure*}[ht]
\centering
\subfigure[Obj01]{\includegraphics[width=5.5cm]{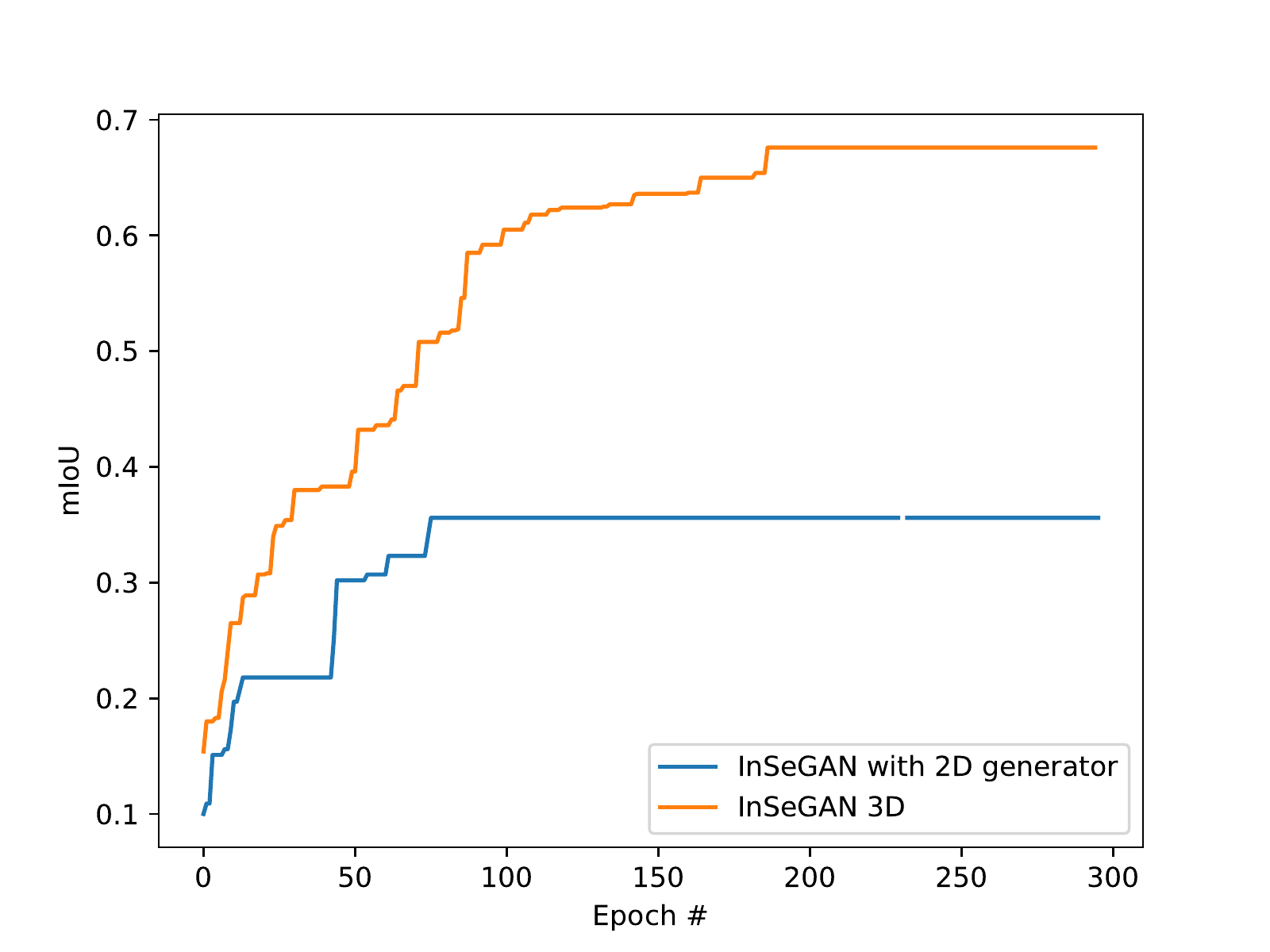}}
\subfigure[Cone]{\includegraphics[width=5.5cm]{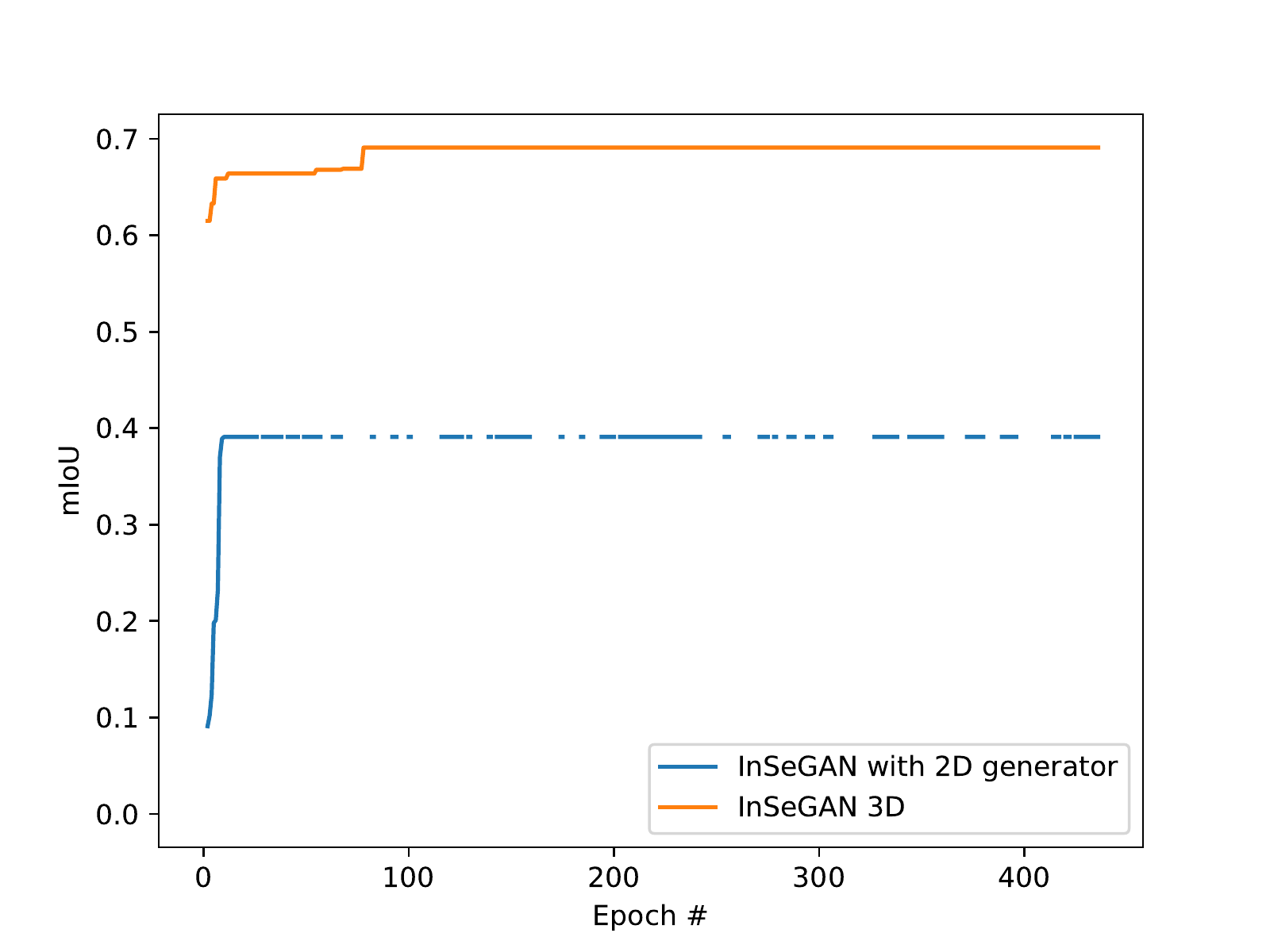}}
\subfigure[Connector]{\includegraphics[width=5.5cm]{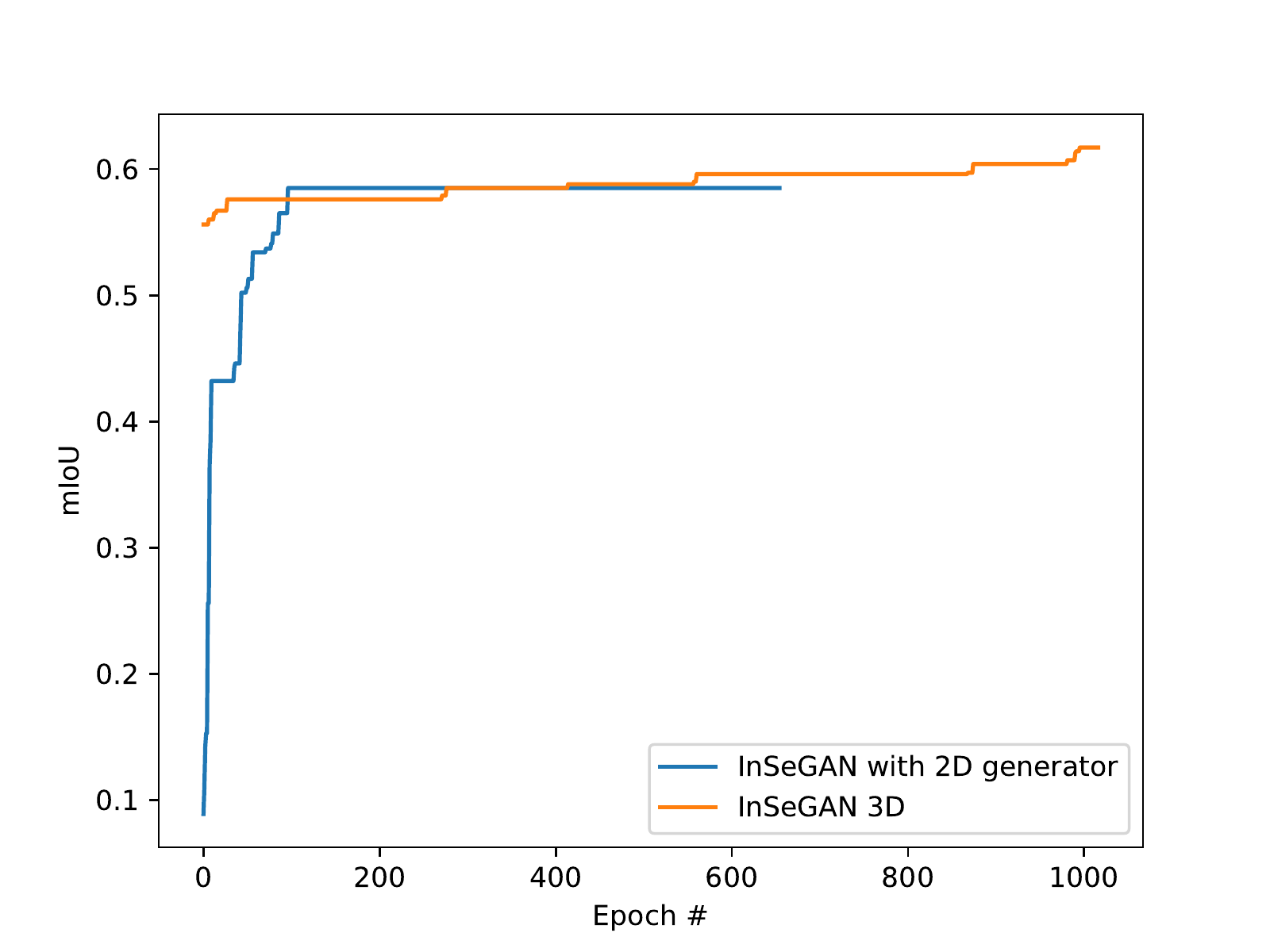}}
\caption{Convergence plots for three objects comparing \name with 3D modules (i.e., using pose encoder, 3D instance template, and STN), shown in orange, with a version in which the 3D modules are replaced by a 2D GAN (i.e., replacing the 3D modules by 2D convolutions and upsampling layers, similar to the encoder and discriminator in reverse). In the figures, we plot mIoU versus the number of training epochs. As is clear, using a 3D GAN leads to better performance and more stable convergence. Note that in the Cone (middle plot), the 2D generator is unstable and often diverges---we reset the optimizer when this happens. This is captured by the discontinuities in the blue plot. In contrast, using the 3D generator leads to very stable training of the generator and discriminator.}
\label{fig:training_performance}
\end{figure*}

\section{Additional Ablative Studies}
In this section, we extend the ablative studies presented in the main paper with additional results, and analyze and substantiate the importance of each choice in \name. 

\vspace{4pt}
\noindent\textbf{Is 3D Generator Important?}
An important choice that we made in \name is the use of a 3D generator instead of a 2D generator. For comparison, we use a standard 2D image-based generator typically used in conditional GANs~\cite{mirza2014conditional}. Specifically, for the 2D generator, we replace the 3D modules in \name (i.e., the 3D implicit template, the pose encoder, and the STN) by 2D convolutions and upsampling layers, similar to those used in the encoder and the discriminator. We perform two experiments to analyze and substantiate our choice: (i) to evaluate the training stability and convergence, and (ii) to evaluate the performance of instance segmentation on the various objects. In Figs.~\ref{fig:training_performance}, we plot the convergence of the 2D and 3D GANs on three objects from our Insta-10 dataset, namely Obj01, Cone, and Connector. We make three observations from these results: (i) 3D GAN is significantly faster than 2D GAN in convergence, (ii) 3D GAN is more stable, and (iii) 3D GAN leads to better mIoU for instance segmentation. In Table 1 of the main paper, we provide comparisons of the 3D and 2D GANs on all the objects in the Insta-10 dataset. Our results show that our 3D generator is significantly better than a 2D generator on a majority of the data classes. 

\noindent\textbf{Do We Need All Training Samples?}
In Fig.~\ref{fig:inc_datasize}, we plot the performance against increasing the number of data samples. That is, we use a random subset of the 10K depth images and evaluate it on our test set. We used subsets with 500, 1000, 3000, 7000, and the full 9800 samples. In Fig.~\ref{fig:inc_datasize}, we plot this performance. As is clear more training data is useful, although this increment appears to be dependent on the object class. In Fig.~\ref{fig:inc_qual_datasize}, we show qualitative results of instance segmentations obtained for different training set sizes to gain insights into what the performances reported in Fig.~\ref{fig:inc_datasize} can be interpreted as. The results show that beyond about 3000 samples, our method seems to start producing qualitatively reasonable instance segmentations, albeit with more data mIoU performance improves.
\begin{figure}
    \centering
     \subfigure[Datasize]{\label{fig:inc_datasize}\includegraphics[width=4cm,trim={0.6cm 0.3cm 1cm 1cm},clip]{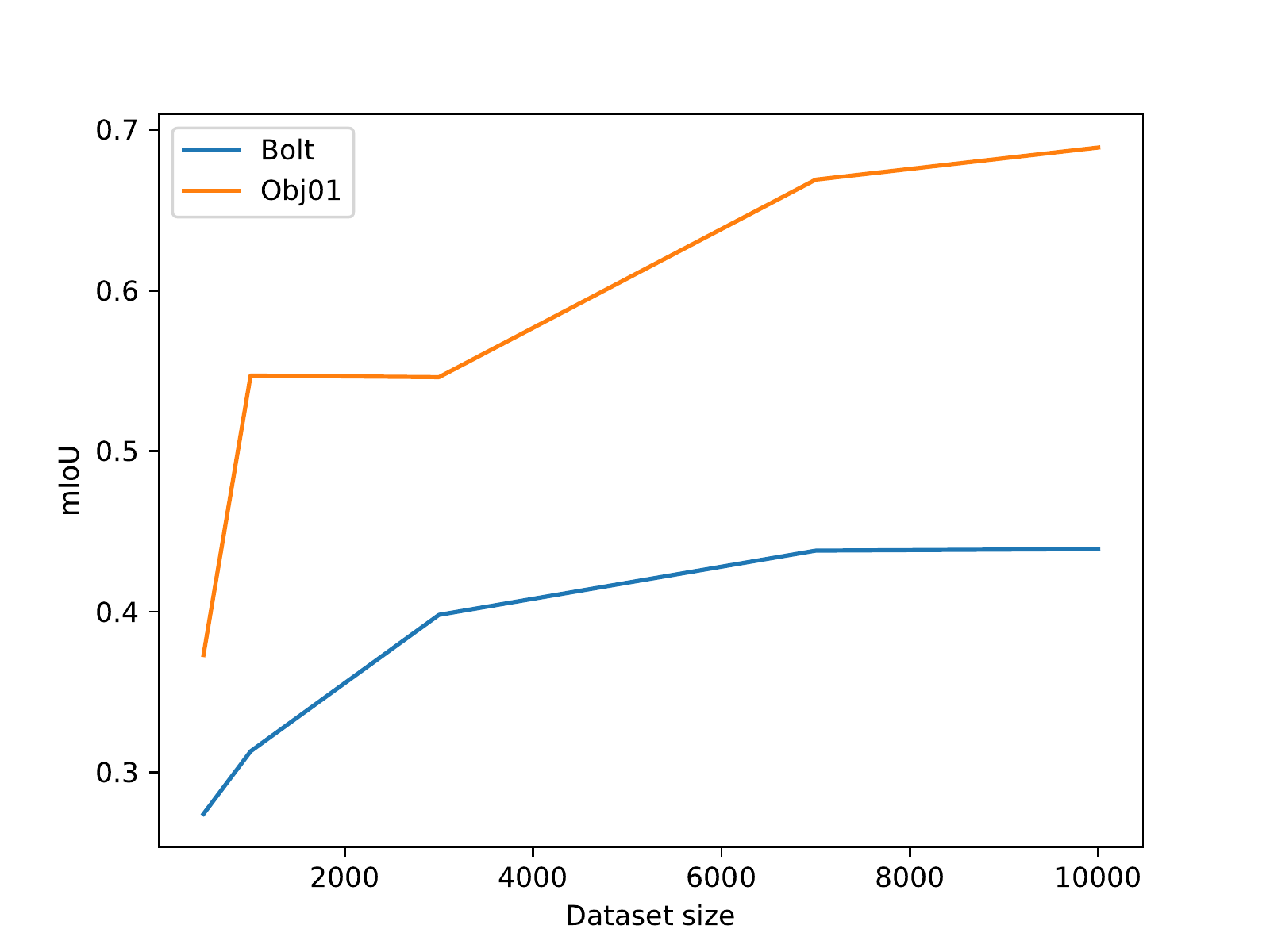}}
    \subfigure[Instances]{\label{fig:inc_instances}\includegraphics[width=4cm,trim={0.6cm 0.3cm 1cm 1cm},clip]{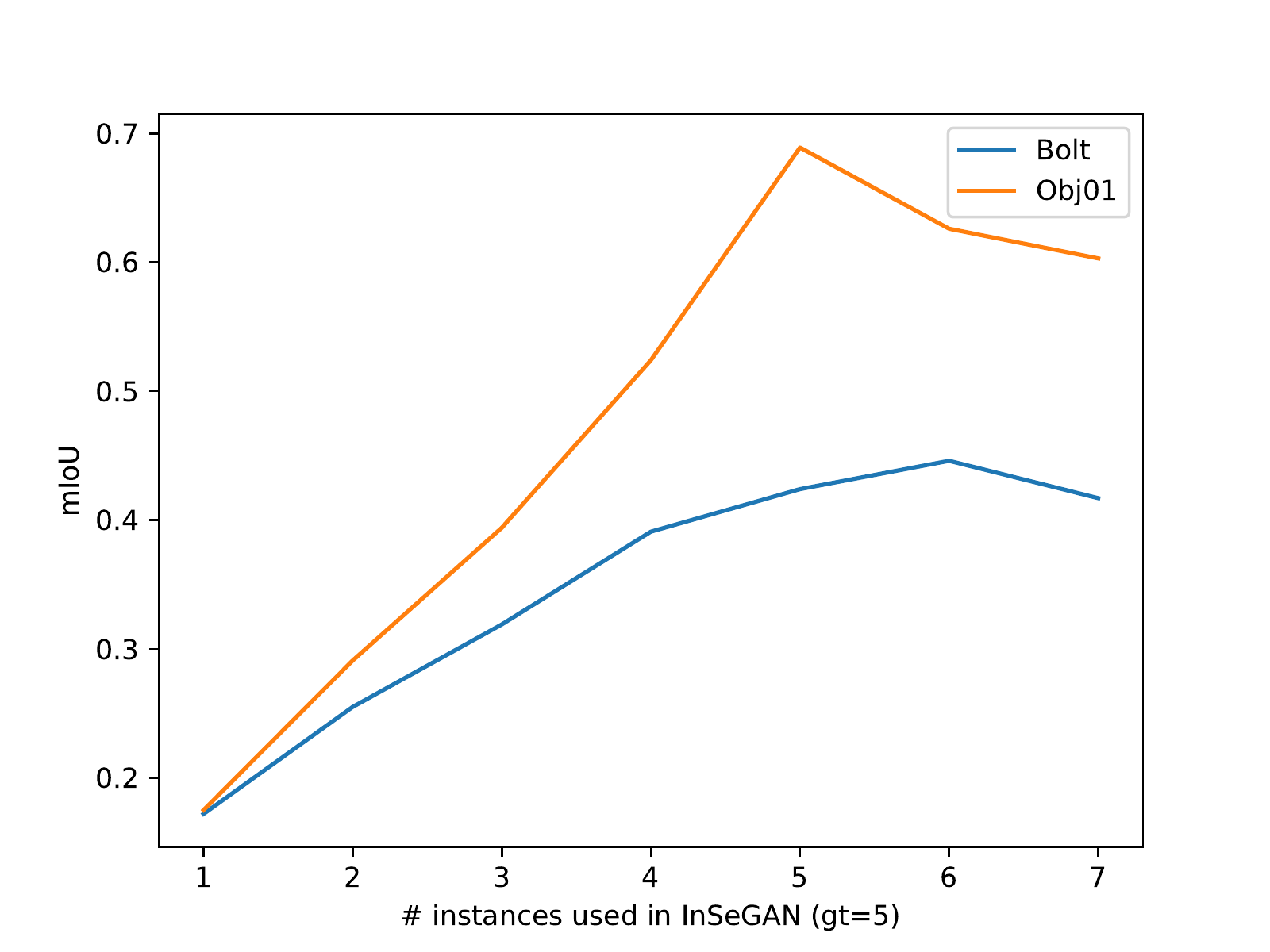}}
    \caption{(a) IoU vs. increasing dataset size. (b) IoU vs. increasing number of instances used in \name ($n$), where the ground-truth number of instances used to generate the data was always $n=5$. Results are shown for two object categories from Insta-10: bolt (blue) and Obj01 (orange).}
\end{figure}

\begin{figure}[h]
\centering
\includegraphics[width=8cm,trim={18cm 1cm 1cm 1cm},clip]{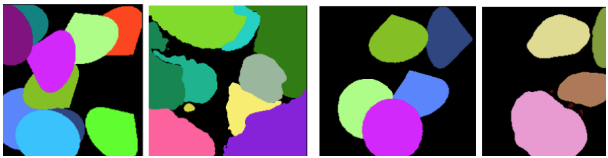}
\caption{Results using Slot Attention~\cite{locatello2020object}.}
\label{fig:sa-attn}
\end{figure}

\begin{figure*}[ht]
    \centering
    \includegraphics[width=8cm,trim={2cm 4cm 4cm 4cm},clip]{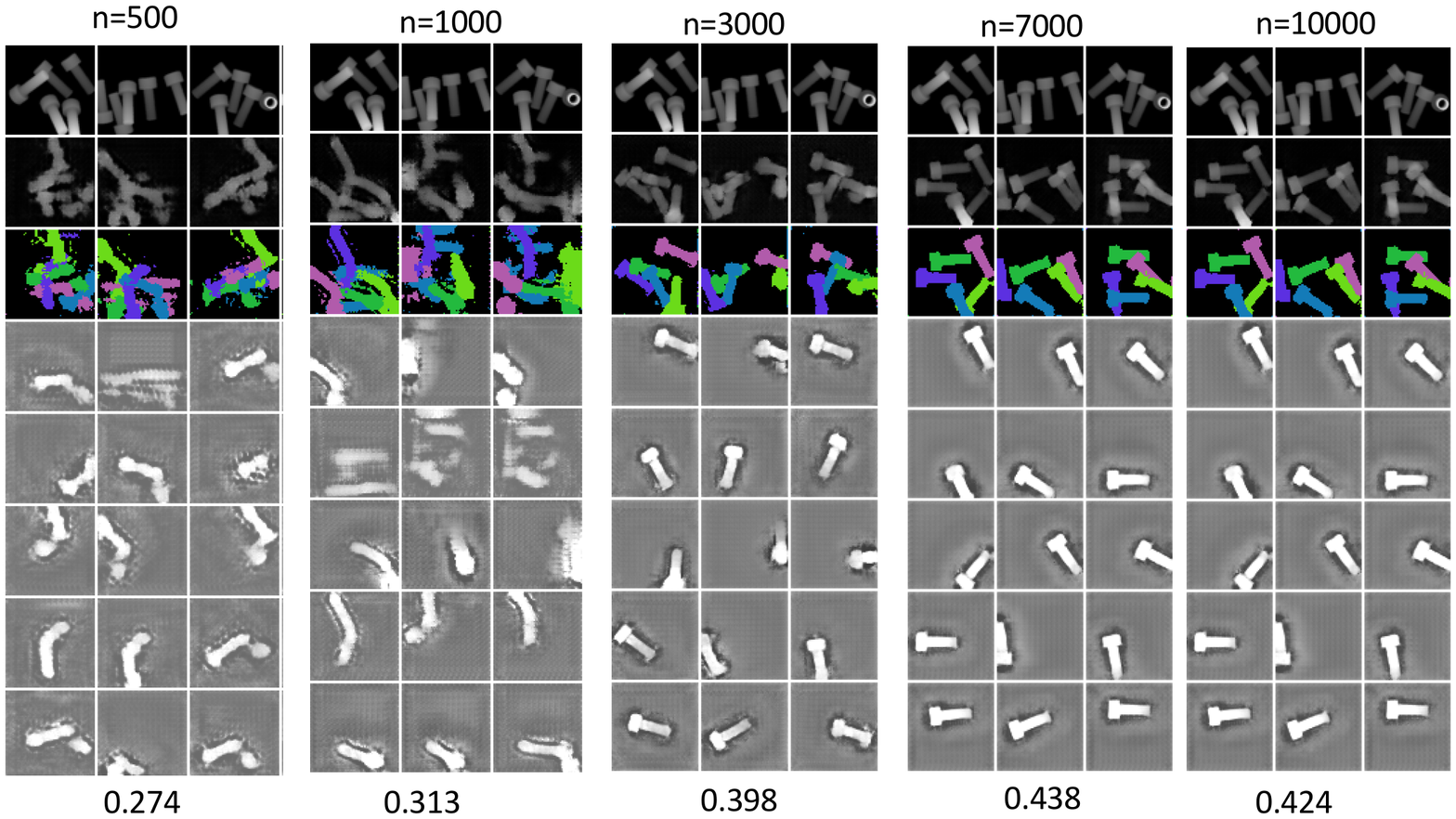}
    \qquad \quad
    \includegraphics[width=8cm,trim={2cm 4cm 4cm 4cm},clip]{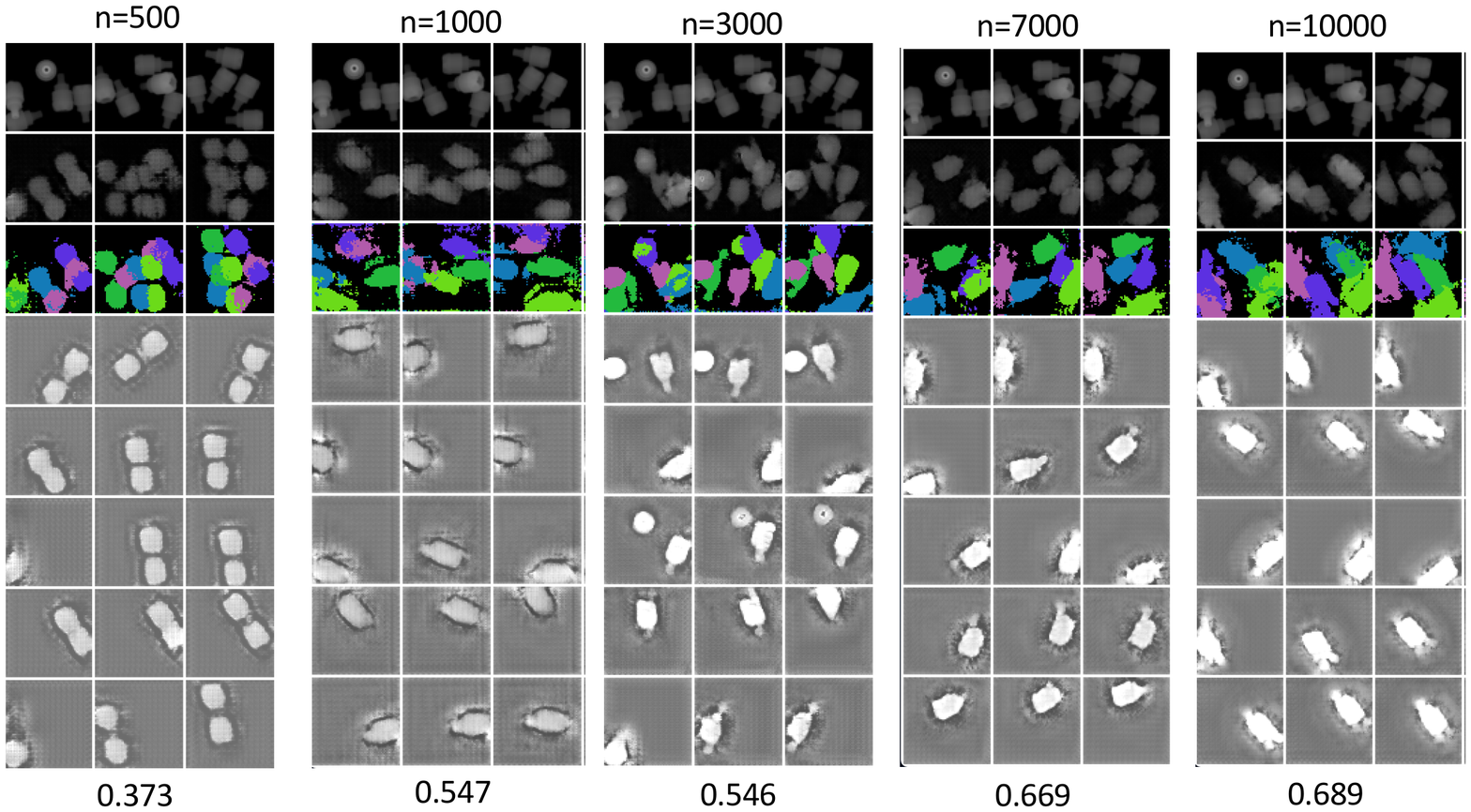}
    \caption{Qualitative instance segmentation results using various data training sizes, for two object classes: Bolt (left) and Obj01 (right). \emph{First row:} input depth image; \emph{second row:} hallucinated depth image by \name; \emph{third row:} inferred instance segmentation; \emph{fourth row onwards:} the single instances hallucinated by \name. The mIoU on the full test set is shown at the bottom.}
    \label{fig:inc_qual_datasize}
\end{figure*}

\begin{figure*}[h]
    \centering
    \includegraphics[width=8cm,trim={2cm 5cm 5.6cm 3cm},clip]{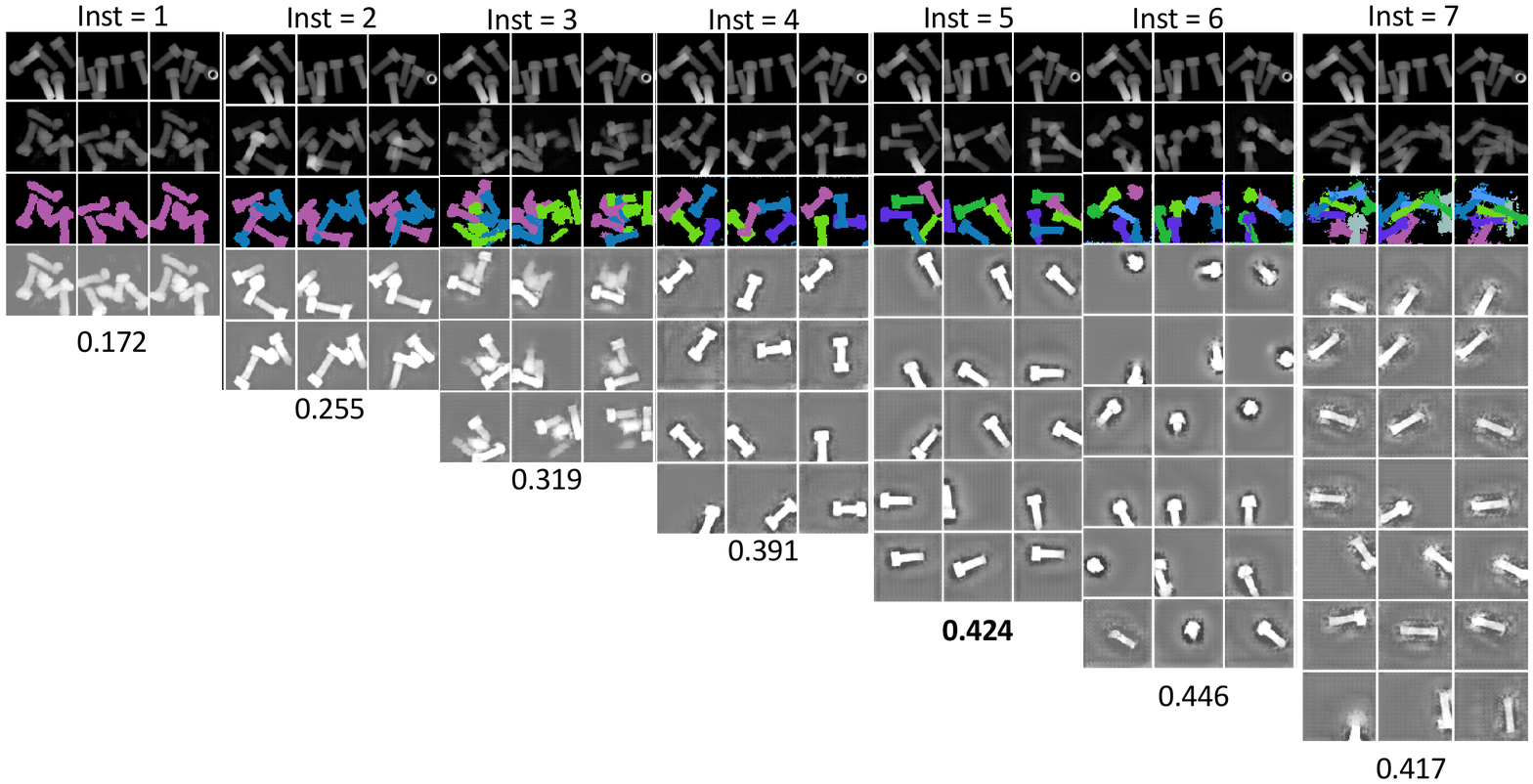}
    \qquad
    \includegraphics[width=8cm,trim={2cm 5cm 5.6cm 4cm},clip]{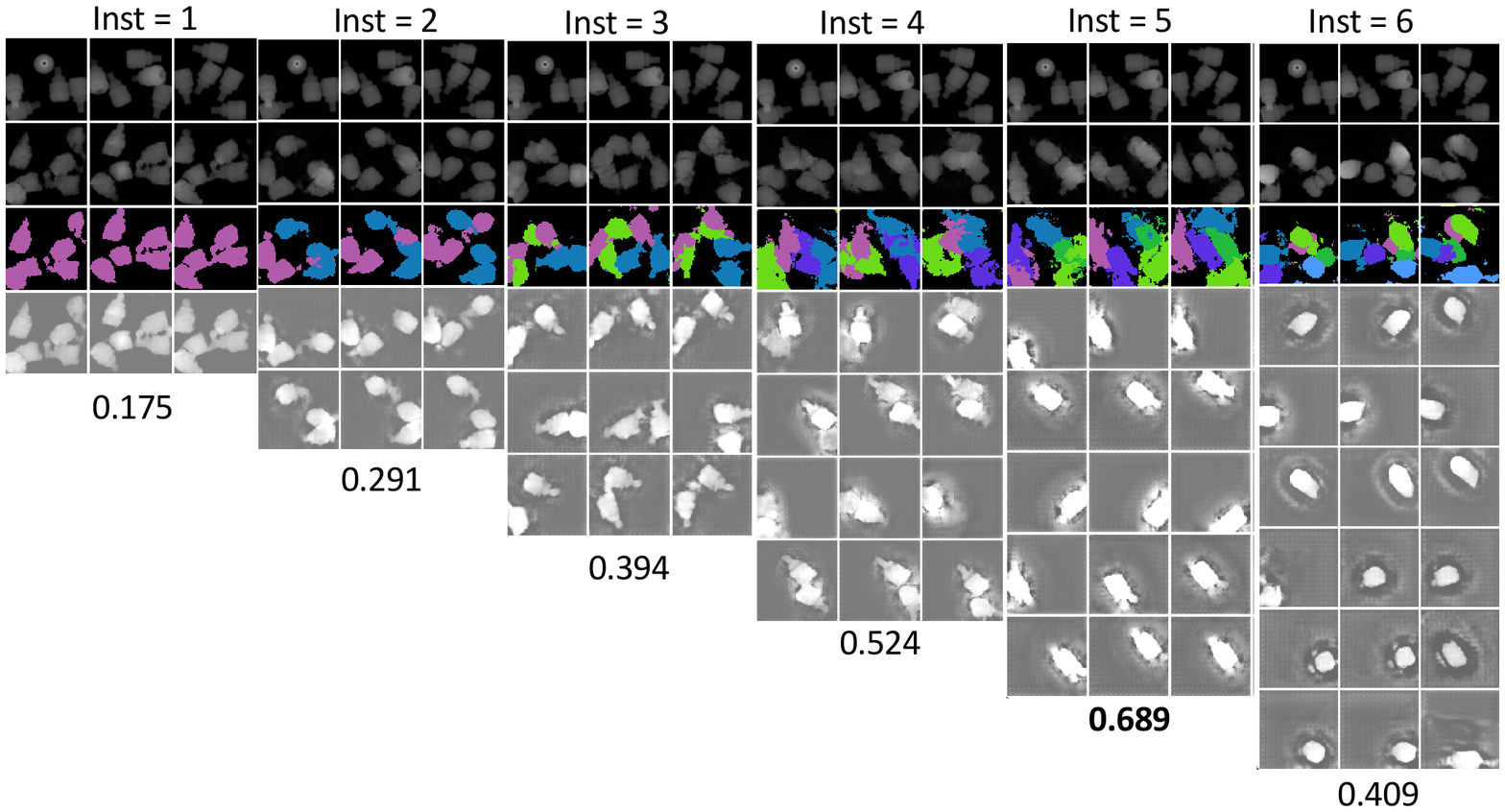}
    \caption{Qualitative instance segmentation results when the number of instances used in \name  is increased, with a fixed ground-truth number of instances ($n=5$ instances). \emph{First row:} input depth image; \emph{second row:} hallucinated depth image by \name; \emph{third row:} inferred instance segmentation; \emph{fourth row onwards:} the single instances hallucinated by \name. The mIoU on the full test set is shown at the bottom.}
    \vspace*{-0.3cm}
    \label{fig:inc_qual_instances}
\end{figure*}

\noindent\textbf{Number of Instances/Disentanglement?}
A key question about our framework is whether the algorithm really needs to know the exact number of instances in order to do well at inference, if the model is trained for a fixed number of instances? What happens if we only have a rough estimate? In this section, we empirically answer this question. In Fig.~\ref{fig:inc_instances}, we plot the performance against increasing the number of instances used in \name; i.e., we increase $n$ from 1 to 7 for the number of noise vectors we sample for the generator. Recall that all our ground-truth depth images consist of 5 instances. The plots in Fig.~\ref{fig:inc_instances} for two objects (Bolt and Obj01) shows that \name performs reasonably well when the number of instances is close to the ground-truth number. In Fig.~\ref{fig:inc_qual_instances}, we plot qualitative results from these choices. Interestingly, we find that using $n=1$ completely fails to capturing the shapes of the objects, while $n=4$ learns a two-sided bolt, and $n=5$ seems to capture the shape perfectly. While $n>5$ seems to show some improvements, it is not consistent across the data classes. Overall, it looks like a rough estimate of the number of instances is sufficient to achieve reasonable instance segmentation performance.

\noindent\textbf{Effect of Noise in the Depth Images?}
In Table~\ref{tab:noise}, we added Gaussian noise $N(0, \sigma)$ to each pixel in the synthetic depth images input to the algorithm for $\sigma=0.1,0.2,0.5$, and pixels depth values in the range $[-1,1]$. We find that \name's performance on noisy depth images is still much better than the performance of K-Means on the noise-free images.
\vspace*{-0.2cm}
\begin{table}[ht]
    \centering
    \begin{tabular}{l|c|c|c|c|c}
    $\sigma$ & KMeans & No noise & 0.1 & 0.2 & 0.5 \\
    \hline
        Bolt & 0.18 &  0.424& 0.352 & 0.326 & 0.318  \\
        obj01 &0.2&0.686& 0.662 & 0.643 & 0.421
    \end{tabular}
    \caption{mIOU for for different noise levels in the depth images.}
    \label{tab:noise}
\end{table}

\subsection{Qualitative Comparisons}
In Fig.~\ref{fig:qual_soa}, we compare qualitative results from \name with those from other methods. For spectral clustering, we used an automatic bandwidth selection scheme in the nearest neighbor kernel construction. For Wu et al.~\cite{wu2019unsupervised}, we use their 1-channel variant, as the 2-channel variant turned out to be very expensive -- it is 32x slower than 1-channel. That said, we did explore the performance of 2-channels on our Bolt class, but did not see any significant performance differences to using 1-channel. We also show comparisons to another recent state of the art method, IODINE~\cite{greff2019multi}. For all of the prior works, we used code provided by the respective authors, and only changed the file path to our dataset. They were trained until convergence (that is, until no change in the objective was found). As is clear from Fig.~\ref{fig:qual_soa}, \name produces more reasonable segmentations than other methods.  In contrast, \name, via modeling the 3D shape of the objects, leads to significant benefits in challenging segmentation settings. In Fig.~\ref{fig:sa-attn}, we show qualitative results using the recent Slot Attention method~\cite{locatello2020object} for the cone class with 5 and 10 instances. 
\begin{figure*}[ht]
\includegraphics[width=16cm,trim={3cm 4cm 9cm 4.5cm},clip]{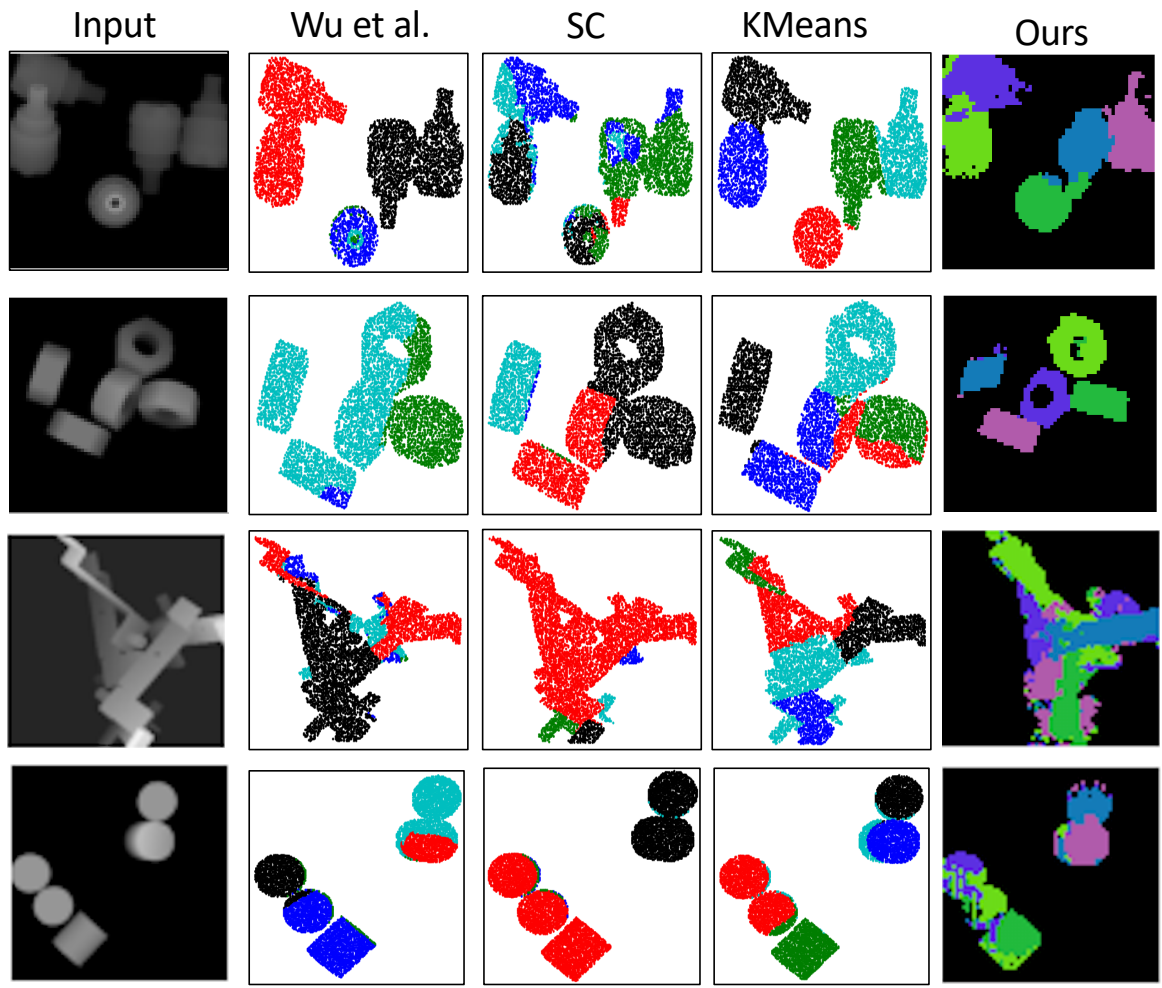}
\caption{Qualitative comparisons of results from \name against other methods. On the right, we show a sample result from a segmentation from the competitive method IODINE~\cite{greff2019multi} (using their code on our data).)}
\label{fig:qual_soa}
\end{figure*}
\subsection{Qualitative Results}
In Figure.~\ref{fig:qual_results}, we show several more qualitative results for each of the 10 object classes in Insta-10. 
\begin{figure*}
    \centering
    \includegraphics[width=7cm,height=4.3cm,trim={0cm 3cm 8.5cm 3cm},clip]{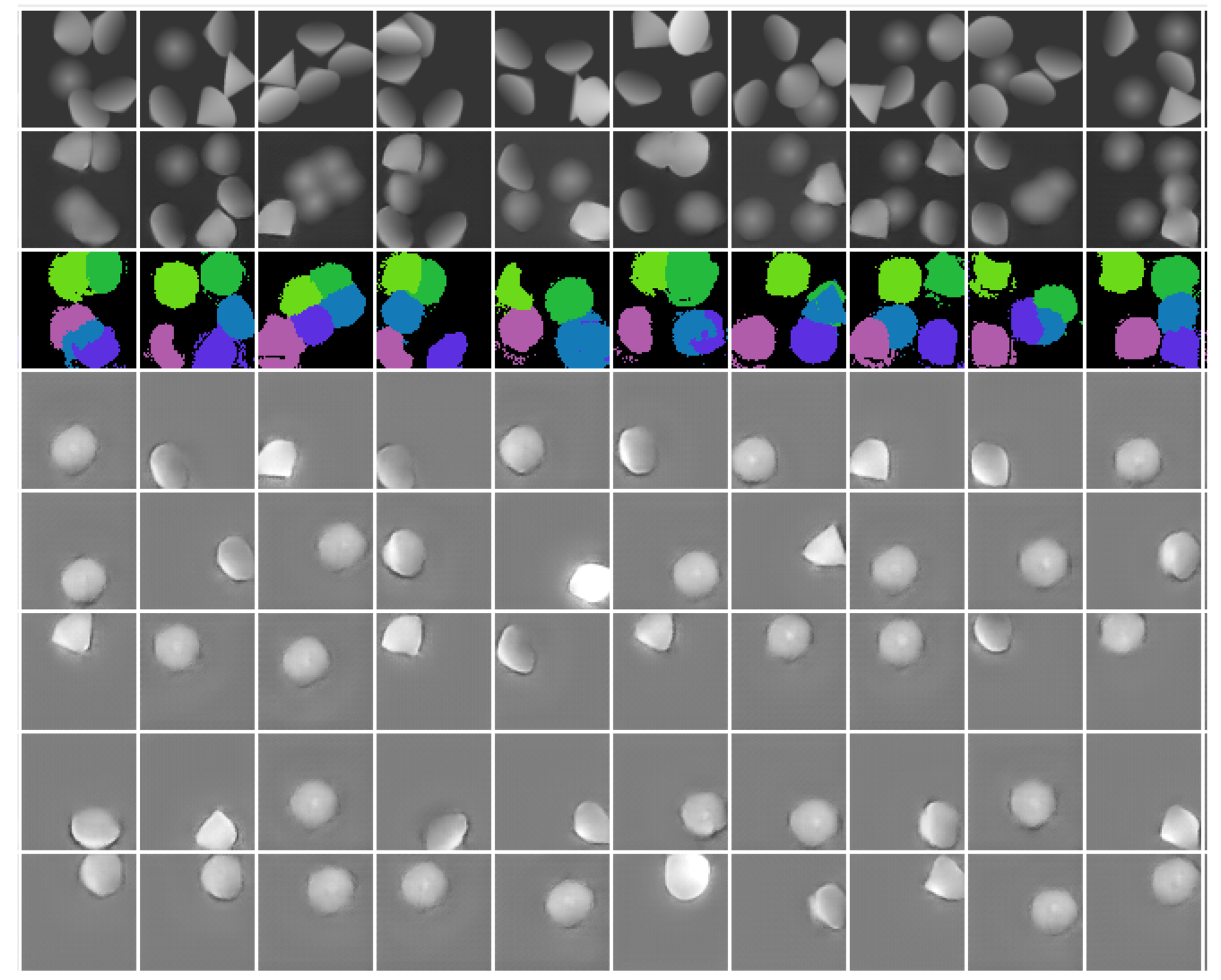}
    \includegraphics[width=7cm,height=4.3cm,trim={0cm 3cm 7cm 3cm},clip]{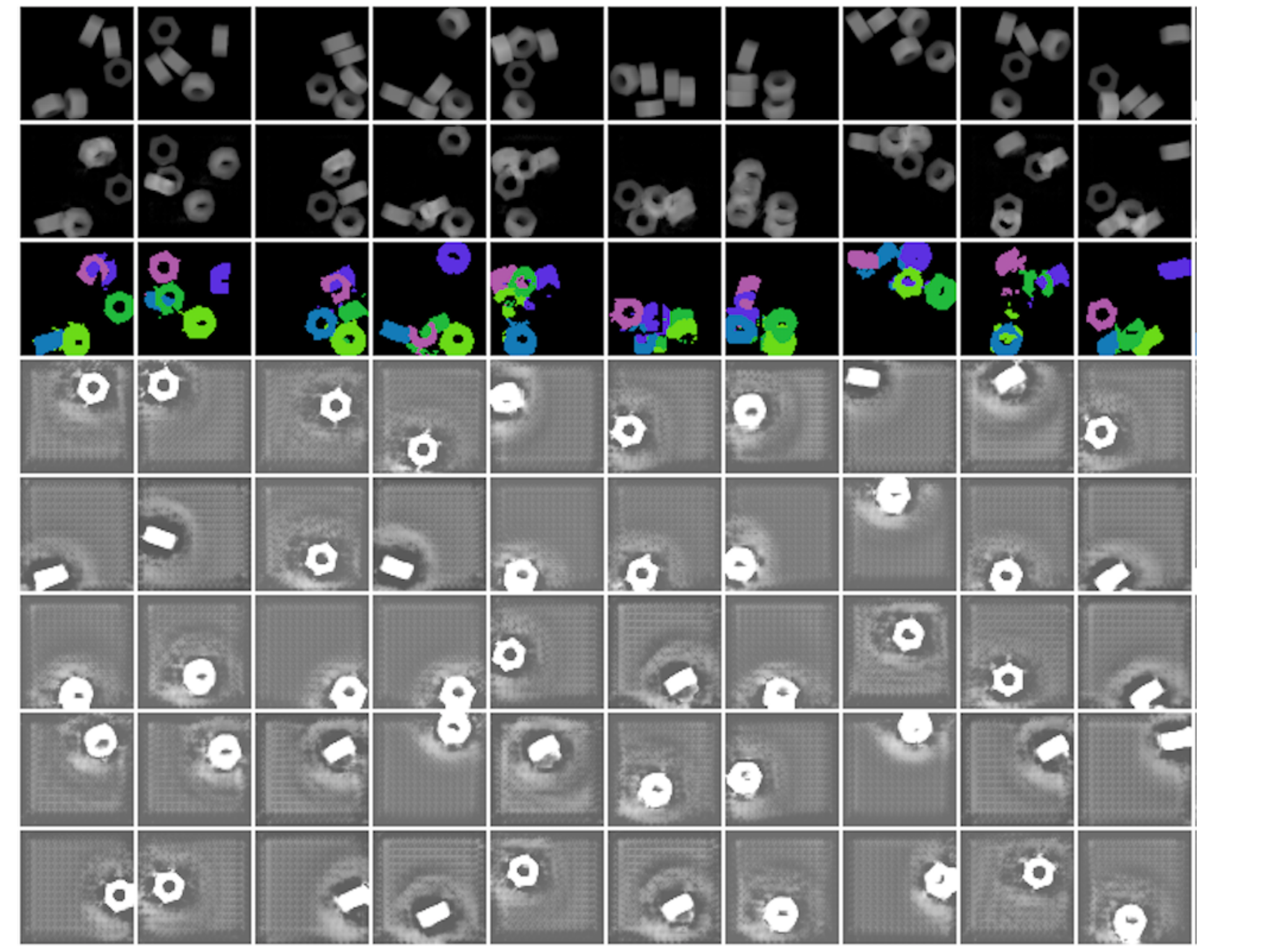}
     \includegraphics[width=7cm,height=4.3cm,trim={0cm 3cm 8.6cm 3cm},clip]{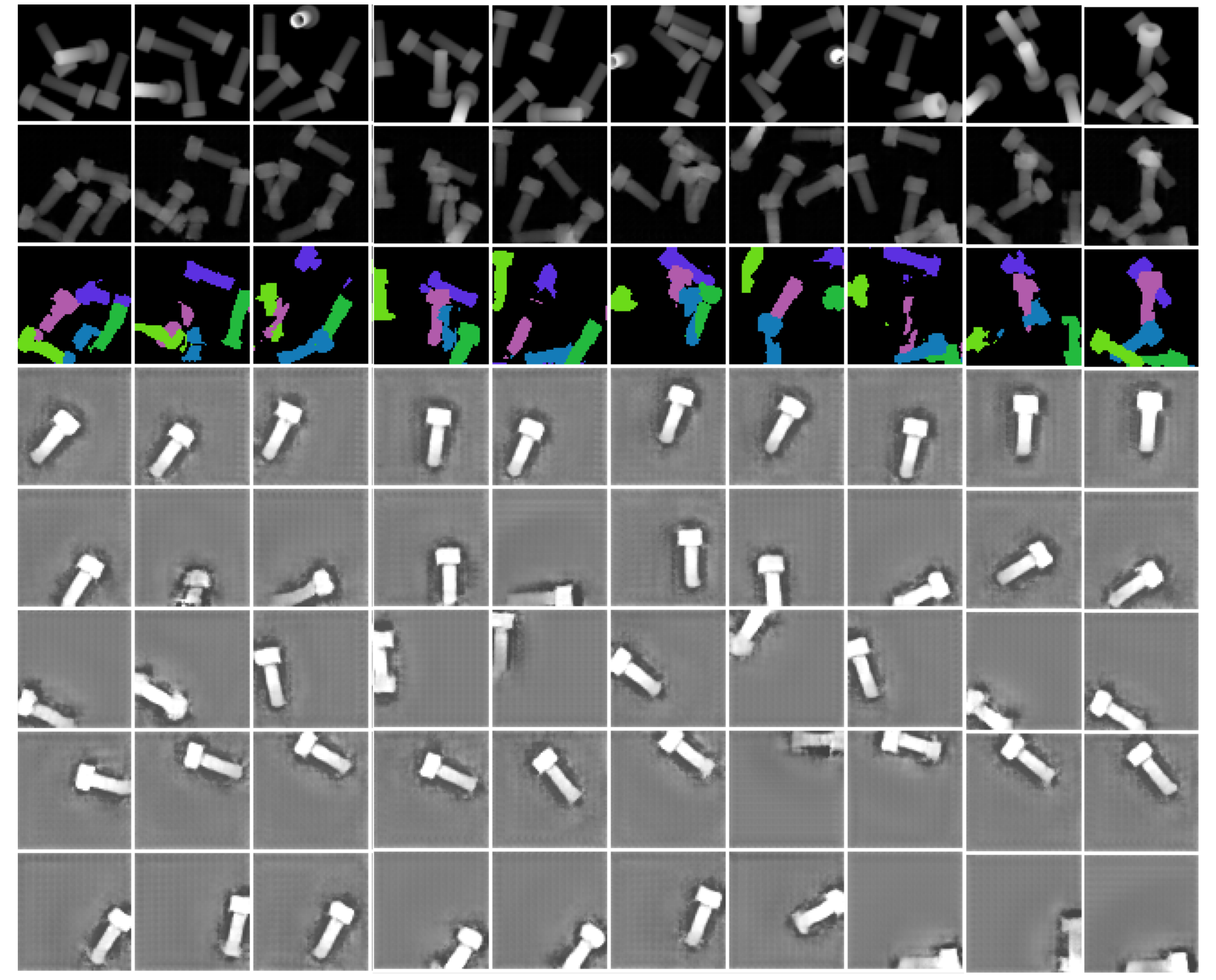}
      \includegraphics[width=7cm,height=4.3cm,trim={0cm 3cm 7.5cm 3cm},clip]{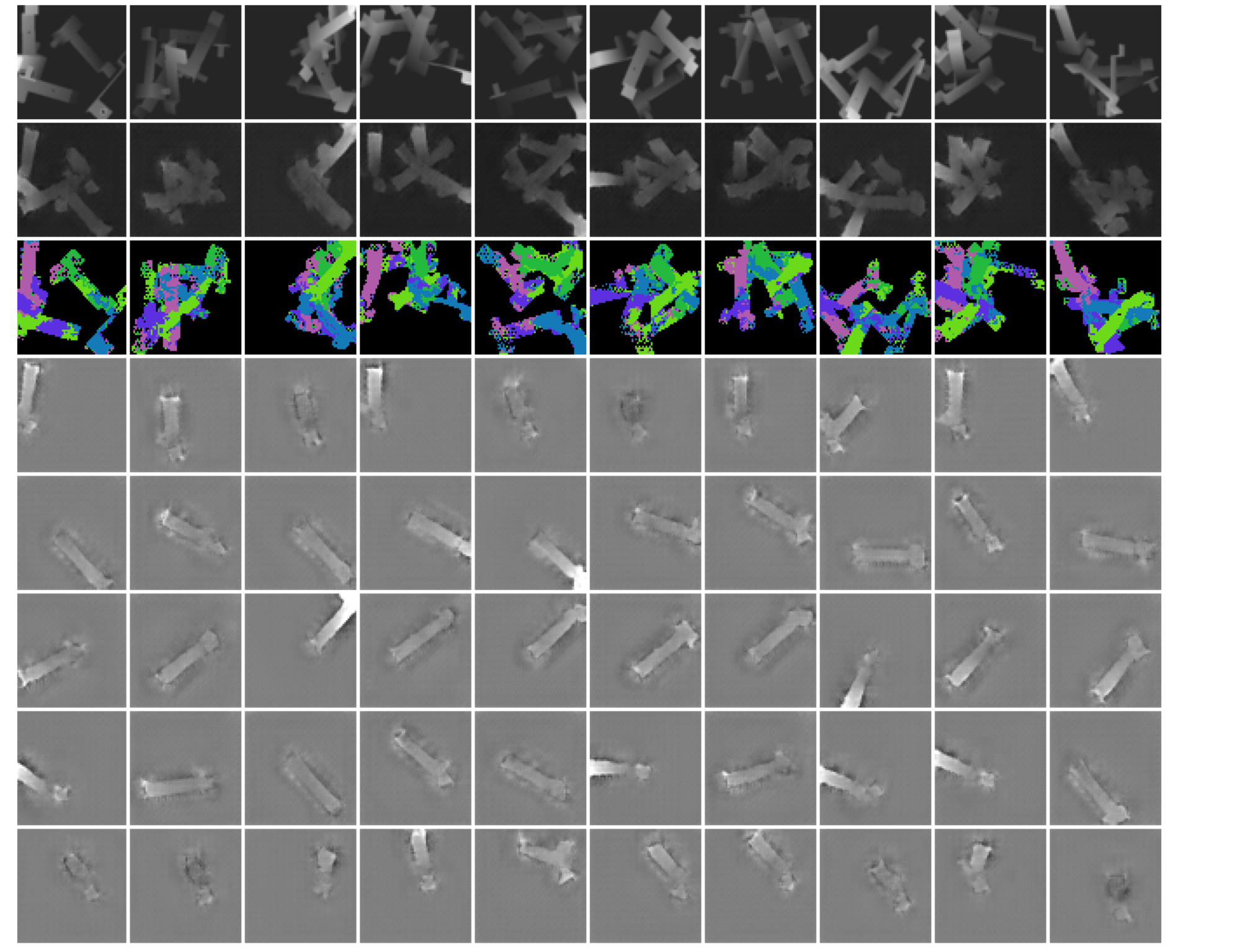}
       \includegraphics[width=7cm,height=4.3cm,trim={0cm 3cm 7.9cm 3cm},clip]{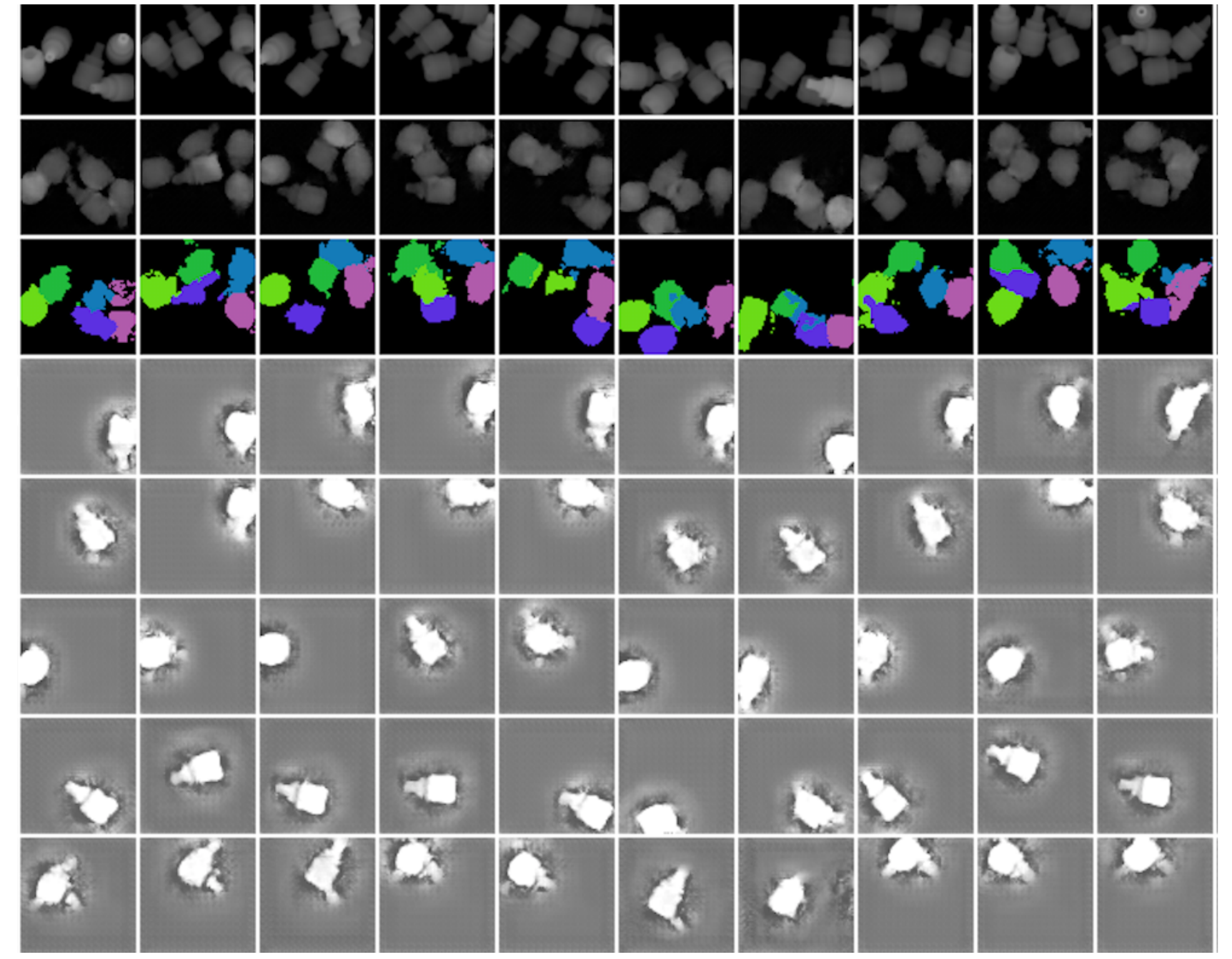}
        \includegraphics[width=7cm,height=4.3cm,trim={0cm 3cm 7cm 3cm},clip]{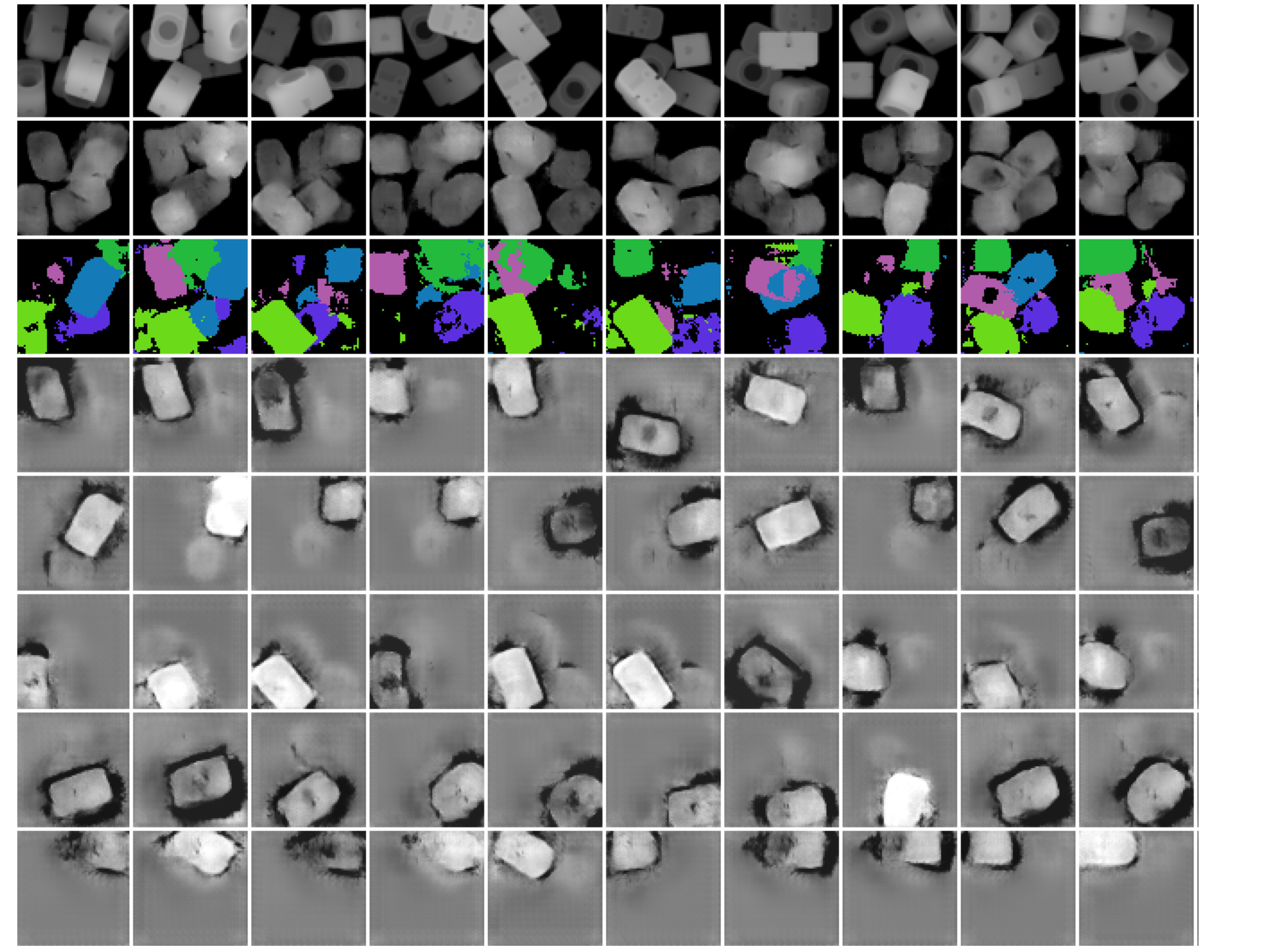}
         \includegraphics[width=7cm,height=4.3cm,trim={0cm 3cm 8.6cm 3cm},clip]{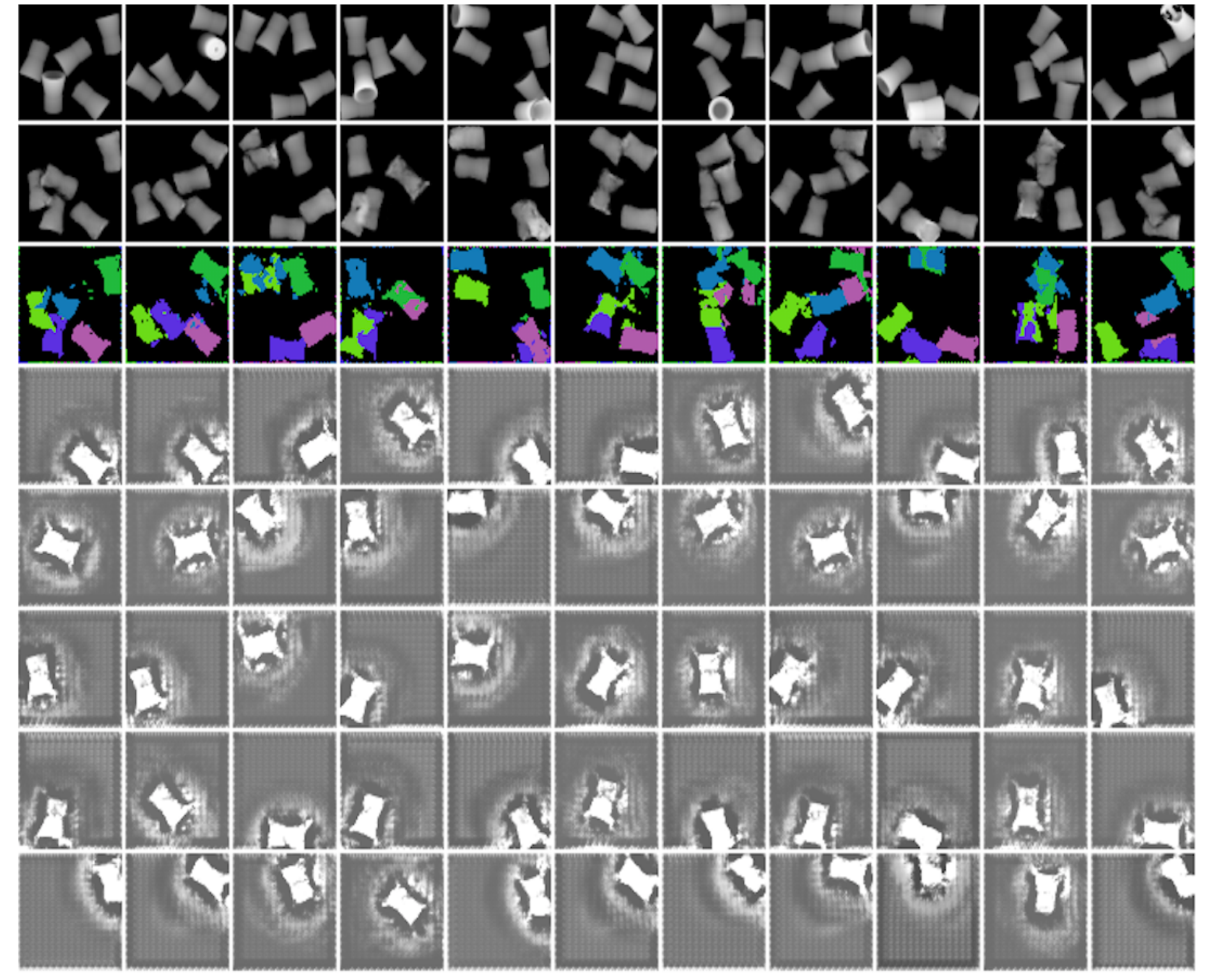}
      \includegraphics[width=7cm,height=4.3cm,trim={0cm 3cm 7.7cm 3cm},clip]{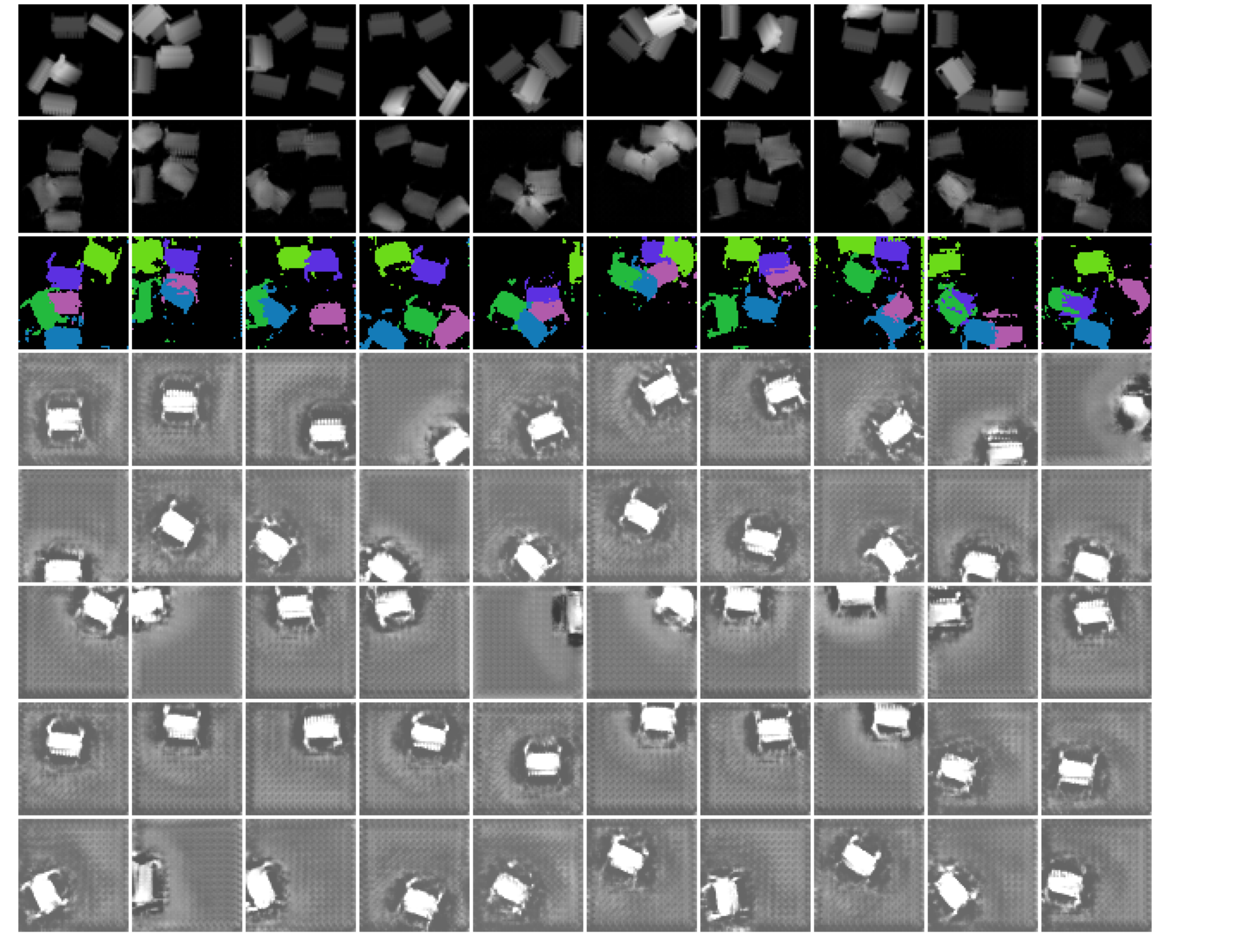}\\
       \includegraphics[width=7cm,height=4.3cm,trim={0cm 3cm 8.5cm 3cm},clip]{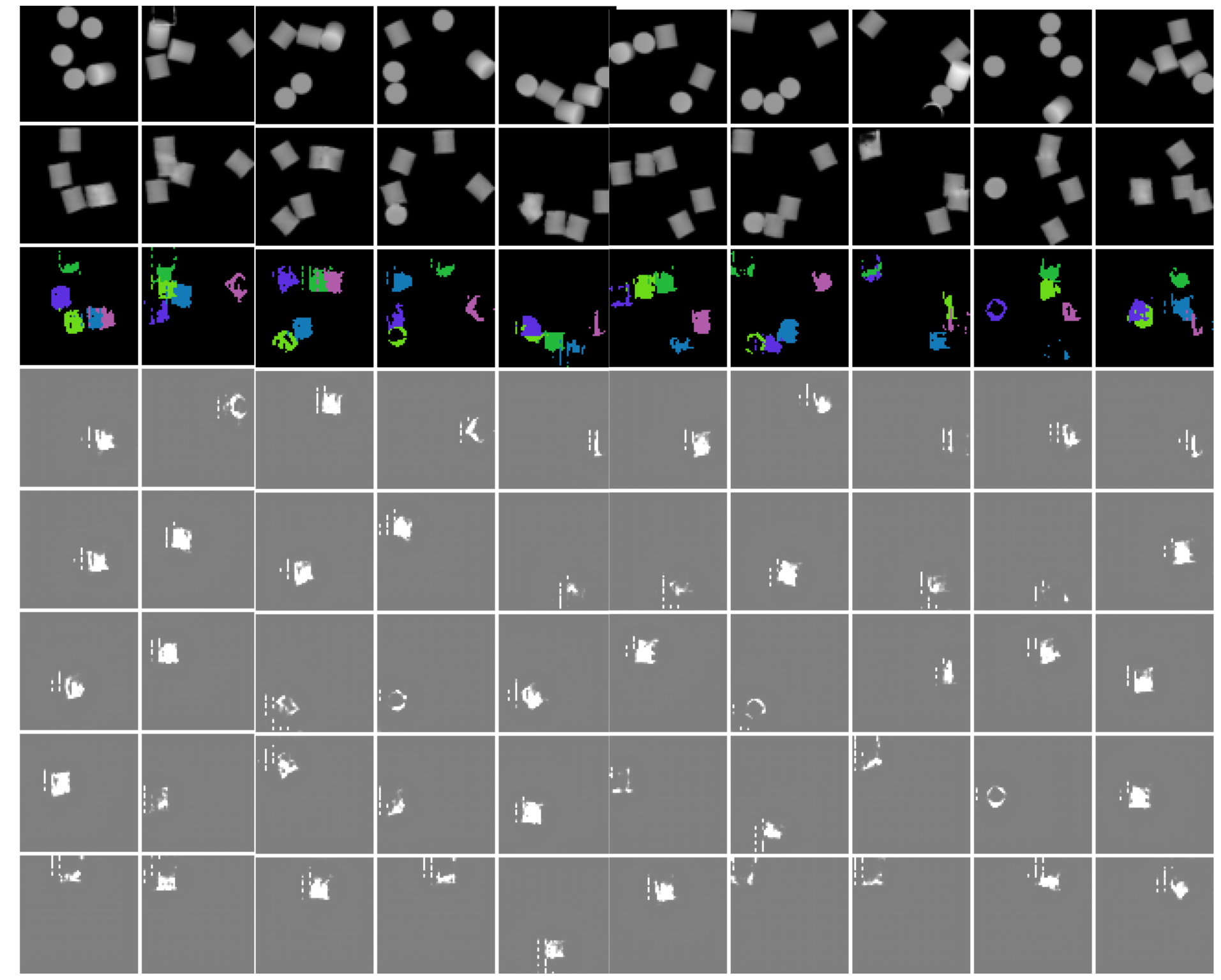}
        \includegraphics[width=7cm,height=4.3cm,trim={0cm 3cm 7.5cm 3cm},clip]{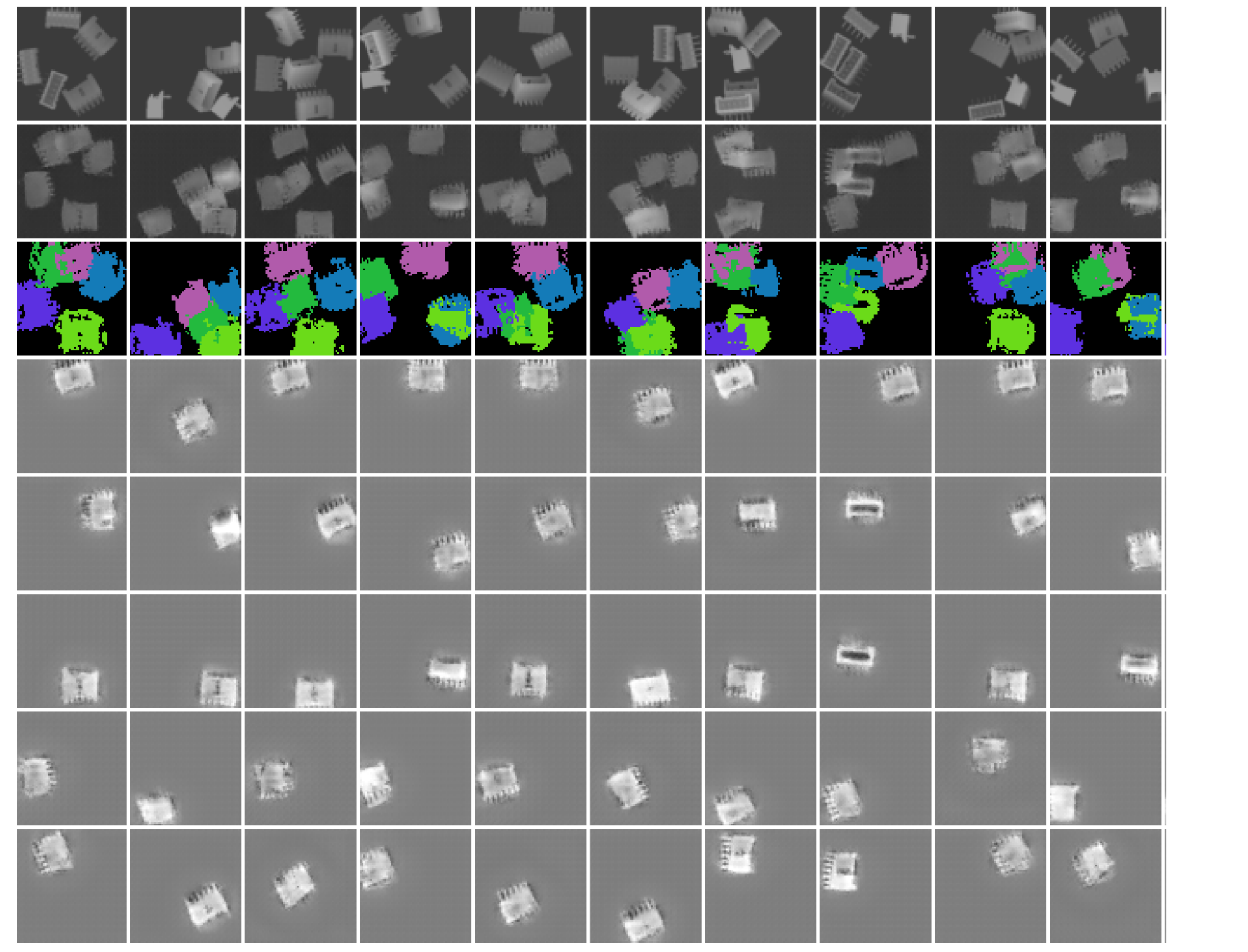}
    \caption{Qualitative results using \name on the 10 object classes in Insta-10. We show 10 segmentation results for each class. \emph{First row:} input depth image; \emph{second row:} hallucinated (reconstructed) depth image by \name; \emph{third row:} inferred instance segmentation; \emph{fourth row onwards:} the single instances hallucinated by \name.}
    \label{fig:qual_results}
\end{figure*}

\end{document}